%% file: main.tex
\documentclass[11pt, letterpaper, logo, onecolumn, copyright, numbering]{minimax}
\usepackage{etoolbox}

\usepackage[authoryear, sort&compress, round]{natbib}

\usepackage[most, breakable, skins]{tcolorbox}
\usepackage{academicons}

\tcbuselibrary{skins}
\usepackage{lipsum}
\usepackage{tabularx}
\usepackage{afterpage}
\usepackage{booktabs}
\usepackage{subcaption}
\usepackage{makecell}
\usepackage{multirow}
\usepackage{multicol} 
\usepackage{array}
\usepackage{float}
\usepackage{listings, listings-rust}
\usepackage{fontawesome5}
\usepackage{amssymb,graphicx}
\usepackage[dvipsnames]{xcolor}
\usepackage{hyperref}
\usepackage{cleveref}
\usepackage{longtable}
\usepackage{graphicx}
\usepackage{pdflscape}
\usepackage{adjustbox}
\usepackage{tikz}
\usetikzlibrary{calc,positioning,chains,shapes,arrows,fit,decorations.pathmorphing,patterns,fadings,shadows,patterns.meta,arrows.meta}
\usepackage{wrapfig}
\usepackage{dialogue}
\usepackage{algorithm}
\usepackage{algorithmic}
\usepackage{colortbl}
\usepackage{mdframed}

\usepackage{listings}

\usepackage{tcolorbox}

\usepackage[dvipsnames]{xcolor}
\usepackage{multicol}

\usepackage{caption}
\input{math_commands.tex}

\input{showcase_format}

\theoremstyle{plain}

\theoremstyle{definition}

\theoremstyle{remark}

\definecolor{medgray55}{gray}{0.55}
\definecolor{medgray}{gray}{0.7}
\definecolor{litegray}{gray}{0.9}
\definecolor{gblue}{RGB}{210, 227, 252}
\definecolor{gred}{RGB}{250, 210, 207}
\definecolor{gyellow}{RGB}{254, 239, 195}
\definecolor{ggreen}{RGB}{206, 234, 214}
\definecolor{gorange}{RGB}{254, 223, 200}

\definecolor{gblue9}{RGB}{23, 78, 166}
\definecolor{gred9}{RGB}{165, 14, 14}
\definecolor{gyellow9}{RGB}{227, 116, 0}
\definecolor{ggreen9}{RGB}{13, 101, 45}
\definecolor{gorange9}{RGB}{176, 96, 0}

\definecolor{myblue}{rgb}{0,0,1}
\definecolor{myred}{rgb}{1,0,0}
\definecolor{mylightgray}{gray}{0.95}

\definecolor{highlightblue}{HTML}{185ABC}

\makeatletter

\renewcommand\paragraph{\@startsection{paragraph}{4}{\z@}%
            {-2.5ex\@plus -1ex \@minus -.25ex}%
            {1.25ex \@plus .25ex}%
            {\itshape\normalsize\bfseries}}
\makeatother
\newcolumntype{L}[1]{>{\raggedright\let\newline\\\arraybackslash\hspace{0pt}}m{#1}}
\newcolumntype{C}[1]{>{\centering}m{#1}}

\newcolumntype{R}[1]{>{\raggedleft\let\newline\\\arraybackslash\hspace{0pt}}m{#1}}

\definecolor{ao}{rgb}{0.0, 0.0, 1.0}

\newcommand\vcent[1]{\vcenter{\hbox{#1}}}

\newcommand\loudspeaker[1][3]{\ensuremath{\vcent{\rule{.6ex}{.6ex}}\kern-.5ex%
  \vcent{\scalebox{.6}[1]{\rotatebox[origin=center]{90}{$\blacktriangle$}}}%
  \ifnum#1>0\relax\kern.05ex\vcent{\scalebox{.4}{\ttfamily)}}%
  \ifnum#1>1\relax\kern-.4ex\vcent{\scalebox{.56}{\ttfamily)}}%
  \ifnum#1>2\relax\kern-.55ex\vcent{\scalebox{.7}{\ttfamily)}}%
  \fi\fi\fi}%
}

\definecolor{green}{rgb}{0.9,0.9,0.9}

\makeatletter
\renewcommand\subparagraph{%
 \@startsection {subparagraph}{5}{\z@ }{3.25ex \@plus 1ex
 \@minus .2ex}{-1em}{\normalfont \normalsize \bfseries }}%
\makeatother

\let\cite\citep

\title{The MiniMax-M2 Series: Mini Activations Unleashing Max Real-World Intelligence}

\reportnumber{}

\renewcommand{\today}{}

\author[*,1]{MiniMax\footnote{Please send correspondence to model@minimax.io.}}

\begin{abstract}
We introduce the MiniMax-M2 series, a family of Mixture-of-Experts language models built around the principle that \emph{mini activations can unleash maximum real-world intelligence}. The flagship M2 contains 229.9B total parameters with only 9.8B activated per token. Designed end-to-end for agentic deployment, the M2 series rests on three components: (i) agent-driven data pipelines producing large-scale, verifiable trajectories across agentic coding and agentic cowork, each grounded in an executable workspace and an artifact-aligned reward; (ii) Forge, a scalable agent-native RL system that adapts to long-horizon agent trajectories, paired with windowed-FIFO scheduling, prefix-tree merging, inference optimization, and a clean training--inference--agent decoupling that supports both white-box and black-box agents; (iii) the latest M2.7 checkpoint takes an early step toward \emph{self-evolution}---autonomously debugging training runs and modifying its own scaffold. Across M2 through M2.7, this combination translates a mini-activation footprint into frontier-tier performance on agentic coding, deep search, office-task, and reasoning benchmarks.
\end{abstract}

\begin{document}

\maketitle

\begin{figure}[H]
\centering
\includegraphics[width=\textwidth]{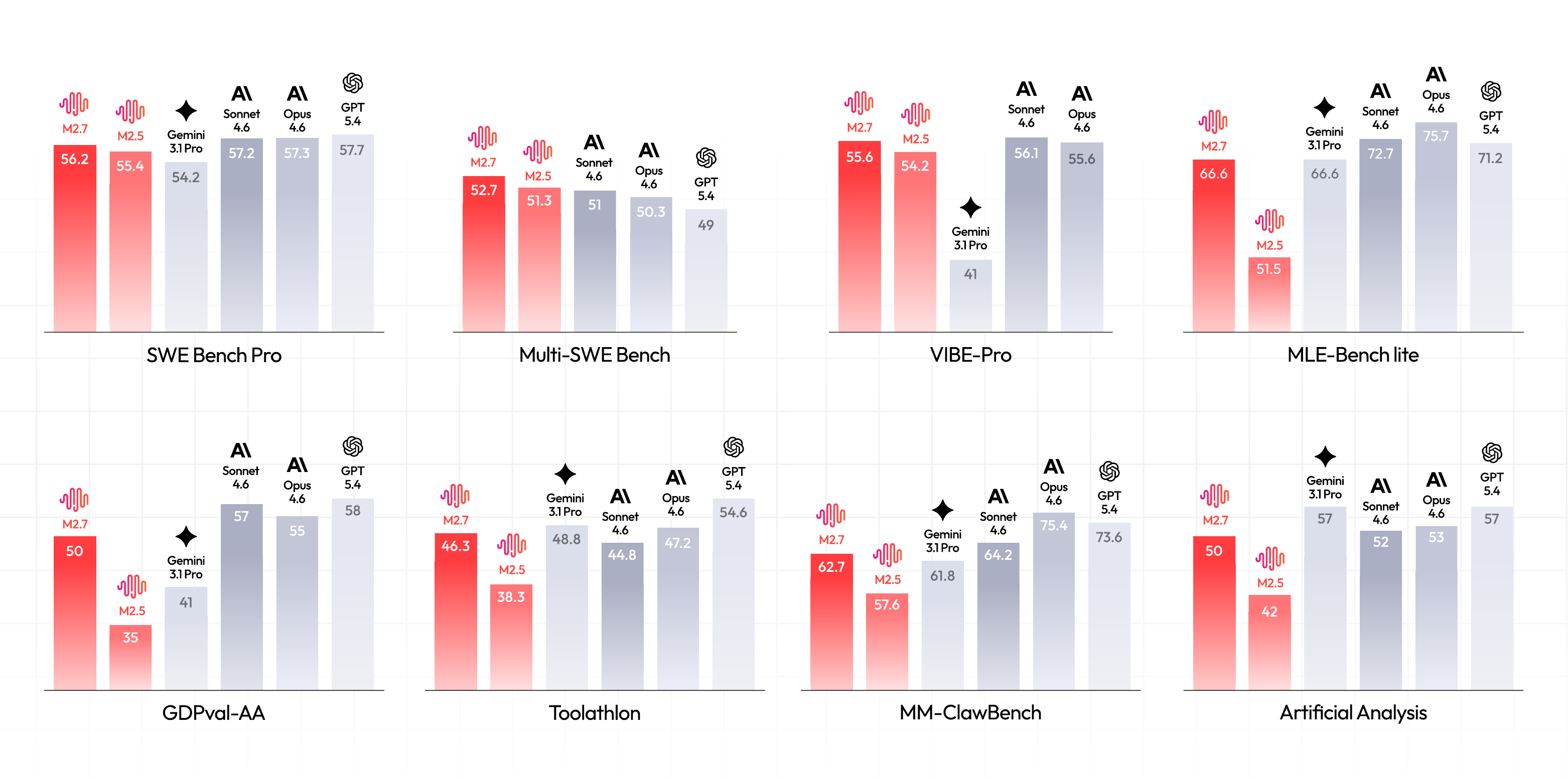}
\caption{Performance of MiniMax-M2.7 versus closed-weight frontier baselines across agentic coding, agentic cowork, and reasoning \& knowledge benchmarks. With only $\sim$10\,B activated parameters, MiniMax-M2.7 remains competitive with substantially larger and more compute-intensive systems.}
\label{fig:main-bar}
\end{figure}
\clearpage

\input{intro}

\input{section/pretrain_architecture}

\input{section/pretrain_data}

\input{section/post_training_data}

\input{section/post_training_sft}

\input{section/post_training_rl}

\input{section/post_training_agentic}

\input{eval}

\input{conclusion}

\bibliography{M2_cite}

\newpage
\input{app}

\end{document}

%% file: math_commands.tex
\usepackage{amsmath,amsfonts,bm}

\def\eqref#1{equation~\ref{#1}}

\def\1{\bm{1}}

\DeclareMathAlphabet{\mathsfit}{\encodingdefault}{\sfdefault}{m}{sl}
\SetMathAlphabet{\mathsfit}{bold}{\encodingdefault}{\sfdefault}{bx}{n}



%% file: showcase_format.tex
\definecolor{ababcol}{HTML}{F14738}
\definecolor{myhailuo2}{HTML}{F97669}
\definecolor{querycol}{HTML}{7964E8}
\definecolor{goldanswercol}{HTML}{FFB43B}
\definecolor{otherscol}{HTML}{FC5BCF}
\definecolor{myhailuo3light}{HTML}{FFA9FA}
\input{colors}
\tcbset{
    showcase/.style={
        fonttitle=\large,
        colback=white!20,  
        colframe=black,   
        coltitle=white,   
        boxrule=0.5mm,    
        arc=2mm,          
        outer arc=2mm,    
        left=1mm,         
        right=1mm,        
        top=1mm,          
        bottom=1mm,       
        width=\textwidth, 
        before skip=0.1pt,
        after skip=0.1pt,
    },
    context/.style={
        fontupper=\scriptsize,
        fonttitle=\large,
        colframe=querycol,     
        coltitle=white,   
        colback=white,    
        boxrule=0.3mm,    
        arc=2mm,          
        outer arc=2mm,    
        left=1mm,         
        right=1mm,        
        top=1mm,          
        bottom=1mm,       
        before skip=1pt,
        after skip=0.1pt, 
    },
    query/.style={
        fontupper=\scriptsize,
        fontlower=\scriptsize,
        colframe=querycol,     
        coltitle=white,   
        colback=white,    
        boxrule=0.1mm,    
        arc=2mm,          
        outer arc=2mm,    
        left=1mm,         
        right=1mm,        
        top=1mm,          
        bottom=1mm,       
        before skip=1pt,
        after skip=0.1pt,
    },
    abab/.style={
        fontupper=\scriptsize,
        fonttitle=,
        colframe=ababcol, 
        coltitle=white,   
        boxrule=0.5mm,    
        arc=2mm,          
        outer arc=2mm,    
        left=1mm,         
        right=1mm,        
        top=1mm,          
        bottom=1mm,       
        width=0.33\textwidth, 
        before skip=0.1pt,
        after skip=0.1pt, 
    },
    others/.style={
        fontupper=\scriptsize,
        colframe=myhailuo3light, 
        coltitle=white,
        boxrule=0.5mm,    
        arc=2mm,          
        outer arc=2mm,    
        left=1mm,         
        right=1mm,        
        top=1mm,          
        bottom=1mm,       
        width=0.33\textwidth, 
        before skip=0.1pt,
        after skip=0.1pt, 
    },
    goldanswer/.style={
        fontupper=\scriptsize,
        colframe=goldanswercol,     
        coltitle=white,   
        boxrule=0.5mm,    
        arc=2mm,          
        outer arc=2mm,    
        left=1mm,         
        right=1mm,        
        top=1mm,          
        bottom=1mm,       
        width=0.33\textwidth, 
        before skip=0.1pt,
        after skip=0.1pt, 
    },
}

%% file: colors.tex
\definecolor{myhailuo1dark}{HTML}{FC8900}
\definecolor{myhailuo2dark}{HTML}{F14738}
\definecolor{myhailuo3dark}{HTML}{D12AAA}
\definecolor{myhailuo4dark}{HTML}{4C4DC2}

\definecolor{myhailuo1}{HTML}{FFB43B}
\definecolor{myhailuo2}{HTML}{F97669}
\definecolor{myhailuo3}{HTML}{FC5BCF}
\definecolor{myhailuo4}{HTML}{7964E8}

\definecolor{myhailuo1light}{HTML}{FFD085}
\definecolor{myhailuo2light}{HTML}{FFA19F}
\definecolor{myhailuo3light}{HTML}{FFA9FA}
\definecolor{myhailuo4light}{HTML}{BDACFB}

\colorlet{myorange}{Orange!20}
\colorlet{mygreen}{LimeGreen!25}
\colorlet{myyellow}{Yellow!30}
\colorlet{myblue}{CornflowerBlue!25}
\colorlet{mybrown}{RawSienna!25}
\colorlet{mypurple}{Orchid!25}
\colorlet{myred}{Red!60}
\colorlet{myorangefull}{YellowOrange!60}
\colorlet{mybrownfull}{RawSienna!60}

\colorlet{myorangethick}{Orange!40}
\colorlet{mygreenthick}{LimeGreen!50}
\colorlet{myyellowthick}{Yellow!60}
\colorlet{mybluethick}{CornflowerBlue!50}

%% file: intro.tex
\section{Introduction}
\label{sec:intro}

Large language models are rapidly migrating from short, single-turn dialogue to long-horizon \emph{agentic} workflows: writing and shipping production code, navigating the open web, operating heterogeneous tools, and producing structured office artifacts across hundreds of interleaved reasoning and action steps~\citep{openai2025gpt5,anthropic2025claude46,google2025gemini31}. This shift exposes two distinct difficulties. First, the inherently ultra-long context of agentic tasks introduces formidable efficiency and cost bottlenecks during both training and inference, particularly under the stringent requirements of large-scale, high-availability production deployment. Second, deployment in the wild demands solving intrinsically complex and high-stakes tasks, such as production-grade software engineering and knowledge-intensive office automation.

To address these twin challenges, we introduce the \textbf{MiniMax-M2 series}, a family of Mixture-of-Experts (MoE) language models built around a single design principle: \emph{mini activations can unleash maximum real-world intelligence}. The flagship M2 is a 62-layer decoder-only Transformer with 229.9B total parameters and only 9.8B activated per token, organized as 256 fine-grained experts~\citep{dai2024deepseekmoe} with sigmoid gating, full multi-head attention with GQA~\citep{ainslie2023gqa}, a 192K-token native context window, and a Multi-Token Prediction (MTP) module~\citep{gloeckle2024better,deepseekai2024v3} that doubles as a speculative-decoding draft path~\citep{leviathan2023fast} at inference. Pre-training on 29.2T tokens establishes the base; the bulk of M2's real-world capability is then constructed by an agent-native post-training pipeline whose components co-evolve from M2 through M2.5 to the latest M2.7.

\noindent \textbf{Main contributions.} The continuous capability evolution and performance enhancements of the MiniMax-M2 series stem primarily from the following technical innovations:
\begin{itemize}
    \item We design high-fidelity, large-scale \textbf{agent data pipelines} tailored for agentic coding, collaborative work (cowork), reasoning, and general knowledge tasks, where each task is accompanied by its corresponding static/runtime environments, verifiable rewards, or credible feedback signals. We find that elevating the reward quality and credibility of each accepted trajectory---whether through executable verification signals or judge-model evidence checking---is of paramount importance to fully unleashing the inherent potential of the base model.

    \item We build \textbf{Forge}, an agent-native RL system engineered for large-scale, general-purpose agentic reinforcement learning, which seamlessly admits both white-box and black-box (API-only) agents within a unified training loop. By decoupling key architectural components---including training, inference, and the agent itself---and pairing this separation with robustness-first algorithmic designs and a meticulous reward system, Forge achieves highly stable RL-time scaling. Furthermore, Forge incorporates windowed-FIFO scheduling to absorb trajectory-length variance, prefix-tree merging, and inference kernels co-designed with our deployment stack, thereby substantially boosting RL training efficiency and scalability.

    \item We demonstrate, in M2.7, an early operational form of \textbf{self-evolution}: the model autonomously triages failed training runs on our own infrastructure, edits its own agent scaffold across tasks and experiments, and is evaluated by running multi-round self-improvement on representative ML-engineering tasks. The within-series gains from M2 $\to$ M2.5 $\to$ M2.7 on agentic benchmarks already reflect this, closing one of the most expensive human-in-the-loop bottlenecks in frontier model development.
\end{itemize}

\noindent \textbf{Results.} Figure~\ref{fig:main-bar} previews the headline numbers for MiniMax-M2.7 across three capability areas. On \emph{agentic coding}, M2.7 reaches 56.2 on \emph{SWE-bench Pro}, 76.5 on \emph{SWE-bench Multilingual}, 52.7 on \emph{Multi-SWE-bench}, and 57.0 on \emph{Terminal-Bench 2.0}. On \emph{agentic cowork}, it reaches 62.7 on \emph{MM Claw}, 77.8 on \emph{BrowseComp}, 50.0 on \emph{GDPval-AA}, and 46.3 on \emph{Toolathlon}. On \emph{reasoning \& knowledge}, M2.7 posts 94.2 on \emph{AIME 2026} and 89.8 on \emph{GPQA-Diamond}. With only $\sim$10\,B activated parameters, MiniMax-M2.7 approaches the performance of the strongest closed-weight frontier systems. We refer the reader to \Cref{sec:eval} for the full benchmark suite, per-benchmark analysis, and within-series progression.

%% file: section/pretrain_architecture.tex
\section{Pre-Training Architecture}

\subsection{Overall Architecture}

M2 is a large-scale sparse language model based on a Mixture-of-Experts (MoE) architecture, designed to scale model capacity while maintaining a low per-token compute budget. It contains 229.9B total parameters, with 9.8B activated per token. The model is implemented as a 62-layer decoder-only Transformer with a hidden dimension of 3{,}072 and a vocabulary size of 200{,}064, and is pre-trained on 29.2T tokens with a maximum context length of 192K.

Each Transformer block in M2 consists of a multi-head self-attention module followed by a Mixture-of-Experts (MoE) feed-forward layer. For attention, M2 adopts full multi-head attention across all layers, using 48 query heads and 8 key-value heads (GQA)~\citep{ainslie2023gqa}. Rotary Position Embeddings (RoPE)~\citep{su2024roformer} are applied throughout the model. This design departs from the hybrid attention mechanisms explored in MiniMax-Text-01~\citep{minimax2025minimax01} and reflects our preference for full attention in large-scale settings (Section~\ref{sec:attention-design}). The MoE feed-forward layer contains 256 fine-grained experts~\citep{dai2024deepseekmoe}, with 8 experts activated per token. Routing is implemented using sigmoid gating with learnable expert-specific bias terms, which improves load balancing while greatly reducing reliance on auxiliary losses~\citep{wang2024auxfree} (Section~\ref{sec:moe}). In addition to the standard next-token prediction objective, we incorporate a Multi-Token Prediction (MTP) module~\citep{gloeckle2024better} during pre-training. This module is expanded during continued pre-training via weight copying to support multi-step speculative decoding~\citep{leviathan2023fast} (Section~\ref{sec:mtp}).

\subsection{Model Design Choice}
\label{sec:design-choice}

M2's design space is dominated by two architectural decisions: how the feed-forward layer is sparsified, and how attention is structured across layers. Each decision was made by deliberately benchmarking against alternatives, with the rationale and supporting evidence detailed below.

\subsubsection{Mixture-of-Experts}
\label{sec:moe}

M2 employs a Mixture-of-Experts (MoE) architecture for its feed-forward layers, with three modifications targeting expressiveness, routing dynamics, and load balancing.

\noindent \textbf{Fine-Grained Experts.} We adopt a fine-grained expert design that uses a larger number of smaller experts, increasing the total expert count while reducing per-expert FFN size. This increases the combinatorial diversity of routing and reduces variance in expert utilization across devices (Table~\ref{tab:fine-grained-experts}).

\begin{table}[h]
\centering
\caption{Ablation studies for the MoE Fine-Grained Experts design and the Multi-Token Prediction (MTP) module, evaluated on \emph{MATH}~\citep{hendrycks2021math}, \emph{MMLU}~\citep{hendrycks2021mmlu}, \emph{ARC-Challenge}~\citep{clark2018arc}, \emph{KorBench}~\citep{ma2025korbenchbenchmarkinglanguagemodels}, and \emph{HumanEval}~\citep{chen2021humaneval}. Bold marks the highest score in each row.}
\label{tab:fine-grained-experts}
\label{tab:mtp-ablation}
\begin{tabular}{lcccc}
\toprule
& \textbf{Shots} & \textbf{Baseline} & \textbf{w/ MTP} & \textbf{w/ Fine-Grained} \\
\midrule
\multicolumn{5}{l}{\textit{Model Configuration}} \\
Activated Params & -- & 2B & 2B & 2B \\
Total Params & -- & 17.8B & 17.8B & 17.8B \\
Training Tokens & -- & 500B & 500B & 500B \\
\texttt{num\_experts} & -- & 32 & 32 & 128 \\
\texttt{topk} & -- & 2 & 2 & 8 \\
\midrule
\multicolumn{5}{l}{\textit{Benchmark Results}} \\
MATH & 4-shot & 19.6 & 21.3 & \textbf{24.1} \\
MMLU & 5-shot & 39.8 & 39.7 & \textbf{40.2} \\
ARC-Challenge & 25-shot & 27.4 & 27.5 & \textbf{27.8} \\
KorBench & 3-shot & 14.1 & \textbf{15.0} & 14.8 \\
HumanEval & 0-shot & 29.7 & 30.1 & \textbf{32.5} \\
\bottomrule
\end{tabular}
\end{table}

\noindent \textbf{Sigmoid Gating.} Instead of softmax-based top-$k$ gating~\citep{shazeer2017moe}, we use sigmoid gating for expert routing. Each expert receives an independent activation score, removing the zero-sum constraint imposed by softmax. This allows multiple experts to be activated simultaneously with high confidence and leads to smoother routing dynamics during training.

\noindent \textbf{Expert Bias.} We introduce learnable bias terms in the gating function as per-expert routing-score shifts. These biases are optimized jointly with model parameters and implicitly regulate expert utilization, allowing the auxiliary load-balancing loss to be greatly reduced.

\subsubsection{Attention}
\label{sec:attention-design}

M2 adopts full multi-head attention across all layers, departing from the hybrid design used in MiniMax-Text-01~\citep{minimax2025minimax01}, which interleaves Lightning Attention~\citep{qin2024lightning} with full attention. Despite the theoretical appeal of efficient attention mechanisms, we found no variant that reliably matches full attention quality in production settings spanning reasoning, coding, and agent tasks.

\noindent \textbf{Evaluation Difficulty.} The core challenge is reliably measuring quality loss. During MiniMax-Text-01 development, our hybrid attention models appeared to match full attention on standard benchmarks (\emph{MMLU}~\citep{hendrycks2021mmlu}, \emph{BBH}~\citep{suzgun2023challenging}, \emph{MATH}~\citep{hendrycks2021math}, \emph{LongBench}~\citep{bai2024longbench}), but at a larger scale, clear deficits emerged in complex multi-hop reasoning. We developed proxy metrics to address these gaps, but the correlation between proxy metrics and real downstream performance is fragile---it may not hold at larger scales or on unseen task distributions. Moreover, the compute required for statistically significant evaluation grows substantially with task complexity, and different architectures interact unpredictably with data distributions and training recipes, making reliable comparisons exceptionally difficult.

\noindent \textbf{Infrastructure Gap.} Linear and sparse attention infrastructure remains less mature than full attention. Many linear architectures are memory-bound even during training. For inference, key challenges remain: sensitivity to low-precision storage, lack of native prefix caching support, and unclear integration with speculative decoding.

\noindent \textbf{Hybrid SWA Experiments.} We extensively explored hybrid Sliding Window Attention~\citep{beltagy2020longformer} variants for M2's attention layers, continuing pre-training for hundreds of billions to trillions of tokens across multiple configurations---varying SWA/full attention ratios, adjusting RoPE settings, exploring intra-layer and inter-layer hybrids, analyzing attention patterns (induction heads~\citep{olsson2022induction}, retrieval heads~\citep{wu2024retrieval}), and adding sink tokens~\citep{xiao2024streamingllm}. During pre-training, all variants showed degraded performance on retrieval, multi-hop reasoning, and in-context learning tasks (Table~\ref{tab:swa-pretrain}). After SFT, the gap became more pronounced specifically at long context: on benchmarks exceeding 32K context (agent tasks and complex long-context evaluations), SWA variants performed significantly worse than full attention. On benchmarks within 32K, differences were mixed and small in absolute terms---SWA matched or even exceeded full attention on some instruction-following and shorter-horizon agent tasks (e.g., \emph{IFBench}, \emph{XBench-ds}), while full attention retained advantages on knowledge-intensive evaluations (e.g., \emph{GPQA-Diamond}, \emph{MMLU-Pro}); see Table~\ref{tab:swa-sft-general}. These findings suggest that hybrid SWA's attention coverage limitations critically impact long-context capabilities while having minimal effect on shorter-context scenarios.

\noindent \textbf{Outlook.} As context lengths grow and GPU compute scaling slows, sub-quadratic attention will become increasingly relevant. We are investing in better long-context data, evaluation methodologies, and infrastructure to enable this transition.

\begin{table}[h]
\centering
\caption{Pretraining evaluation at the M2 architecture scale: full attention baseline vs.\ hybrid SWA, covering general knowledge (\emph{MMLU}~\citep{hendrycks2021mmlu}, \emph{MATH}~\citep{hendrycks2021math}), long-context retrieval (\emph{HELMET}~\citep{yen2024helmet}, \emph{RULER}~\citep{hsieh2024ruler}), and in-context translation (\emph{MTOB}~\citep{tanzer2024mtob}). Bold marks the highest score in each row.}
\label{tab:swa-pretrain}
\begin{tabular}{lcc}
\toprule
& \textbf{Baseline} & \textbf{w/ SWA} \\
\midrule
HELMET ICL & \textbf{75.8} & 72.7 \\
MMLU & 85.5 & 85.6 \\
MATH & 60.3 & 60.3 \\
RULER 128K CWE & \textbf{90.0} & 72.0 \\
RULER 128K MQ & \textbf{99.0} & 93.0 \\
RULER 32K CWE & 99.0 & 99.0 \\
RULER 32K MQ & 99.0 & 99.0 \\
MTOB K-e Bleurt & \textbf{60.0} & 45.0 \\
MTOB e-k ChrF & \textbf{44.8} & 27.2 \\
\bottomrule
\end{tabular}
\end{table}

\begin{table}[h]
\centering
\caption{SFT benchmark at the M2 architecture scale: full attention baseline vs.\ hybrid SWA, covering general reasoning/knowledge and agentic tasks. General benchmarks: \emph{AIME 2025}~\citep{aime2025}, \emph{ARC-AGI-1}~\citep{chollet2019arcagi}, \emph{GPQA-Diamond}~\citep{rein2024gpqa}, \emph{MMLU-Pro}~\citep{wang2024mmlu}, \emph{IFBench}~\citep{pyatkin2025ifbench}. Agent benchmarks: \emph{SWE-verified}~\citep{jimenez2024swebench}, \emph{Terminal-Bench}~\citep{merrill2026terminalbench}, \emph{BrowseComp-zh}~\citep{zhou2025browsecompzhbenchmarkingwebbrowsing}, \emph{GAIA-103}~\citep{mialon2024gaia}, \emph{XBench-ds}~\citep{chen2025xbench}, \emph{$\tau^2$-Bench}~\citep{barres2025tau2benchevaluatingconversationalagents}. Bold marks the highest score in each row.}
\label{tab:swa-sft-general}
\begin{tabular}{lcc}
\toprule
& \textbf{Baseline} & \textbf{w/ SWA} \\
\midrule
\multicolumn{3}{l}{\textit{General Benchmarks}} \\
AIME 2025 & 86.7 & 86.7 \\
ARC-AGI-1 & 38.9 & \textbf{39.6} \\
GPQA-Diamond & \textbf{75.3} & 72.7 \\
MMLU-Pro & \textbf{80.5} & 80.1 \\
IFBench & 23.1 & \textbf{27.2} \\
\midrule
\multicolumn{3}{l}{\textit{Agent Benchmarks}} \\
SWE-verified & \textbf{54.7} & 50.2 \\
Terminal-Bench & \textbf{26.7} & 23.8 \\
BrowseComp-zh & \textbf{32.8} & 28.7 \\
GAIA-103 & \textbf{53.4} & 51.5 \\
XBench-ds & 58.0 & \textbf{63.0} \\
$\tau^2$-Bench retail & 62.3 & \textbf{67.5} \\
$\tau^2$-Bench telecom & \textbf{32.5} & 21.0 \\
\bottomrule
\end{tabular}
\end{table}

\subsection{Multi-Token Prediction}
\label{sec:mtp}

M2 incorporates Multi-Token Prediction (MTP)~\citep{gloeckle2024better}, which trains the model to predict the next $K$ tokens jointly. This design provides richer training signals and enables speculative decoding~\citep{leviathan2023fast} at inference time.

\begin{figure}[h]
\centering
\includegraphics[width=0.85\textwidth]{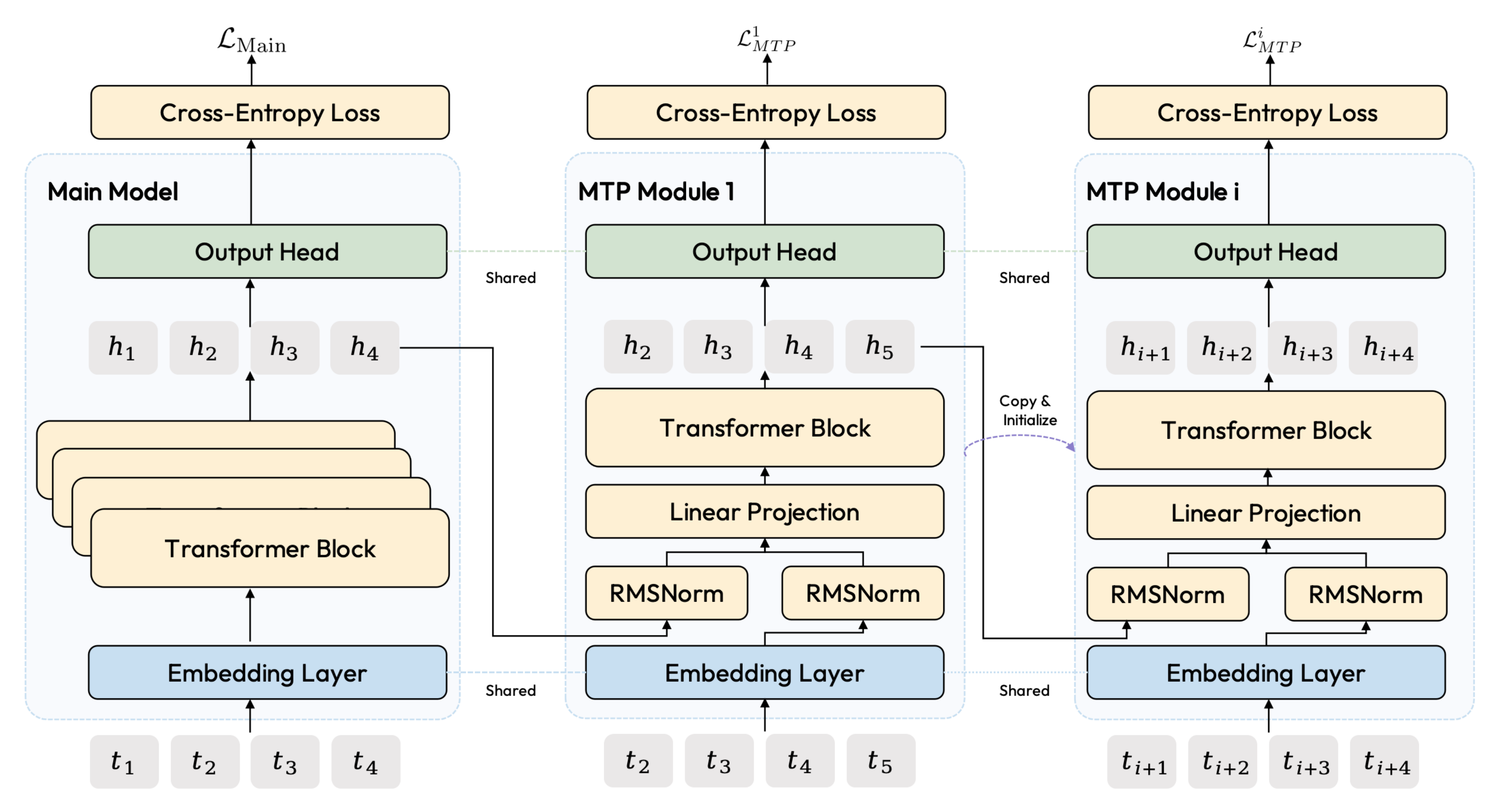}
\caption{Multi-Token Prediction (MTP) module architecture used in M2.}
\label{fig:mtp}
\end{figure}

\noindent \textbf{Pre-training Stage.} During pre-training, M2 is trained with a single MTP module ($K=1$) following the design of DeepSeek-V3~\citep{deepseekai2024v3} (Figure~\ref{fig:mtp}), with an initial MTP loss weight of 0.3, which is annealed to 0.1 during the decay phase. As shown in Table~\ref{tab:mtp-ablation}, our ablation indicates that MTP consistently improves model performance across benchmarks, with the largest gains on reasoning-heavy tasks.

\noindent \textbf{Expansion via Weight Copying.} To support multi-step speculative decoding, we expand from one to three MTP modules ($K=3$) during the decay phase of continued pre-training. Rather than random initialization, we copy weights from the main model to initialize the MTP modules. This strategy is critical for two reasons: (1) copy-initialized modules converge significantly faster than randomly initialized ones, which otherwise start with high loss and temporarily degrade the main model; (2) it minimizes disruption to the main model representations during the transition. After expansion, we first freeze the main model and train only the MTP modules for a short period until their loss stabilizes, then switch to joint training of all modules. We also explored keeping the main model frozen throughout, but found that the MTP modules converged to a worse final quality under this MTP-only schedule than under joint training.

\noindent \textbf{Inference.} At inference time, the three MTP modules generate draft tokens that are verified by the main model in a single forward pass, providing throughput improvement while maintaining identical output quality to standard autoregressive decoding.

%% file: section/pretrain_data.tex
\section{Pre-Training Data}

\noindent \textbf{Training Data.} The pre-training corpus encompasses a comprehensive and meticulously curated dataset, incorporating diverse sources including web documents, academic literature, books, programming code, and structured question-answering content. We employ a combination of model-based reward scoring and auxiliary classifiers to assess document quality across multiple dimensions, and apply a balanced sampling strategy that upweights high-quality content while retaining sufficient category diversity.

\noindent \textbf{Data Distribution.} The pre-training data mixture is carefully balanced across domains, with code, mathematics, and STEM content significantly upsampled relative to their natural distribution. The remaining portion consists of general web content, books, and other domain-specific data, ensuring broad coverage of world knowledge and linguistic diversity. During the constant phase of pre-training, we train on a total of 19.9T tokens.

\noindent \textbf{Long-Context Extension.} Following the initial pre-training phase, we adopt a multi-stage training procedure to progressively extend the model's context window from 8K tokens through 32K and ultimately to 192K tokens. The decay phase uses a total data budget of 9.3T tokens, comprising both short-text decay data and long-context data, where high-quality code concatenation, naturally long-form PDF documents, and thematically related document packing serve as the primary sources of long-context training samples. During the decay phase, we mix in high-quality data to consolidate the model's capabilities while extending its effective context length.

%% file: section/post_training_data.tex
\section{Post-Training Data Collection}
\label{sec:post-training-data}

\subsection{Agentic Coding}
\label{sec:agentic-coding-data}

We collect post-training data for agentic coding across three complementary domains: software engineering (SWE), application development (AppDev), and terminal interaction tasks, covering repository-level code evolution, full-stack development, and interactive terminal environments.


\begin{figure}[h]
\centering
\includegraphics[width=\textwidth]{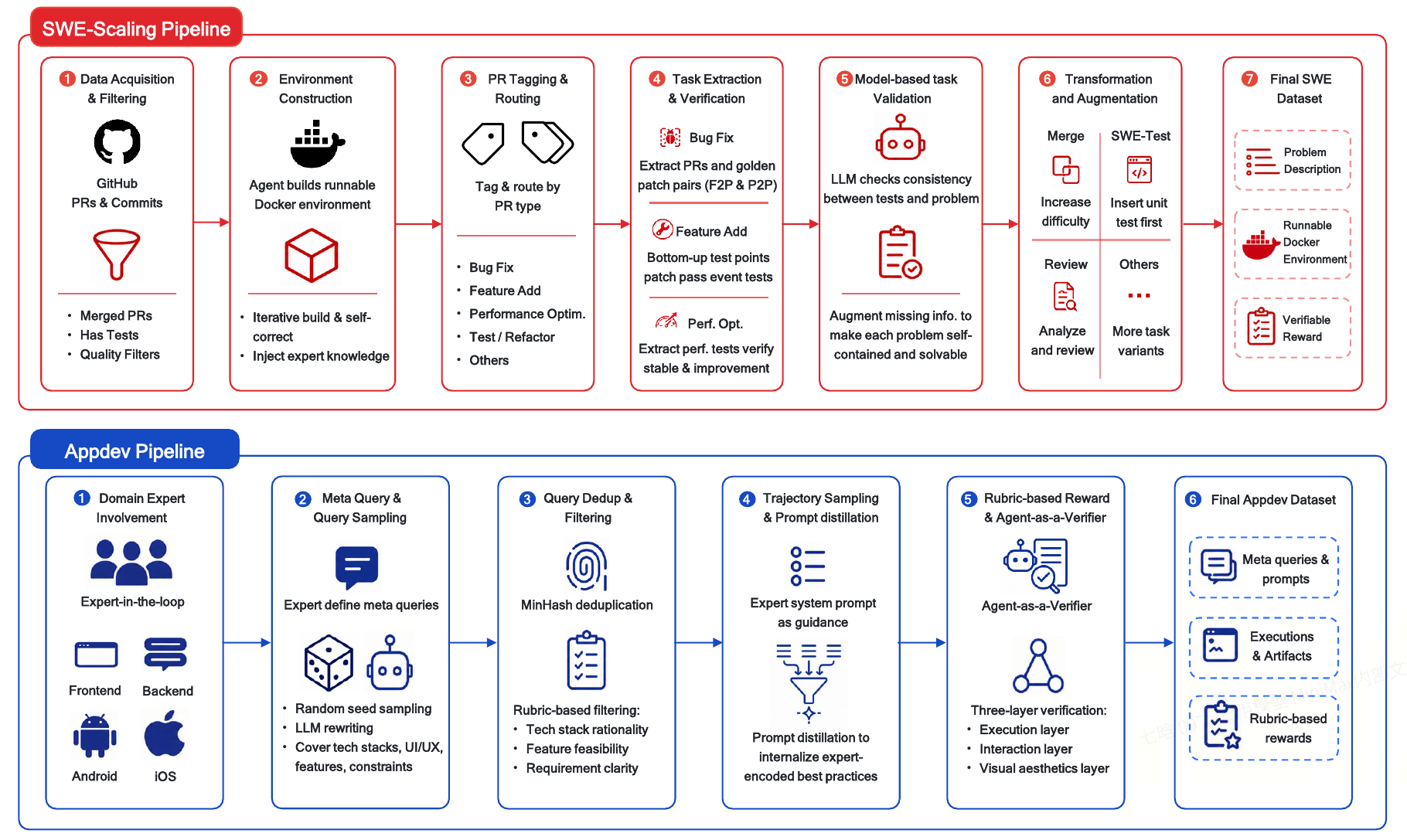}
\caption{The agentic coding data pipelines for SWE and AppDev tasks.}
\label{fig:swe-appdev}
\end{figure}

\subsubsection{Real-Data Driven Collection: Software Engineering Tasks}

Constructing training data for coding agents poses three coupled challenges: achieving broad task diversity, ensuring objective verifiability, and scaling to the volumes that large-scale training demands. GitHub serves as a rich and naturally structured source for collecting such data: a well-structured pull request captures a description, associated code changes, and test cases that provide objective correctness signals. However, raw PR data is inherently noisy and cannot be used directly, motivating our construction of a real-data driven SWE-scaling pipeline: an agent-driven automated data pipeline based on raw GitHub data to produce diverse, verifiable SWE-style datasets and environments. Specifically, the pipeline proceeds through the following six consecutive stages.

\begin{itemize}
    \item \textbf{PR collection and filtering.} The first stage of the pipeline involves large-scale crawling of public GitHub repositories with permissive licenses to collect pull requests and their linked issues, which together provide code diffs, test files, and problem statements. Since raw GitHub data is inherently noisy, we apply a rule-based quality filter on the PRs that were eventually merged, along with additional criteria such as the presence of relevant test cases.

    \item \textbf{Agent-synthesized multi-language Docker environments.} In the SWE-scaling pipeline, we aim to construct a runnable Docker environment for each PR.
    However, we observe that environment synthesis is less reliable in non-Python settings due to heterogeneous dependencies and version conflicts. To address this, we introduce an agent-driven execution loop that incorporates expert knowledge, enabling iterative generation and refinement of build scripts guided by execution feedback. The key dimensions we address are as follows:

    \begin{itemize}
        \item \textit{Build system orchestration in compiled languages.} Compiled languages such as Java, Go, Rust, and C++ require complex toolchain coordination, including compiler versions, build tools, and dependency resolution.
    
        \item \textit{Heterogeneous execution and testing interfaces.} Different languages expose distinct build and testing pipelines, requiring unified yet adaptable execution interfaces for environment setup and validation.
    
        \item \textit{Repository-level structural variability.} Differences in project organization and dependency specification across repositories necessitate adaptive strategies for code localization and test execution.
    \end{itemize}

    \item \textbf{PR tagging and task diversification.} After constructing the Docker environment for each PR, we perform PR-level tagging and routing. GitHub PRs span a broad taxonomy of task types, including bug fixes, feature additions, performance optimizations, refactoring, and test construction. Such routing is necessary because different task types require distinct formulations of downstream verifiable rewards.

    \item \textbf{Test-based verifiable reward construction.} We design task-specific reward functions grounded in test-case execution, as different PR types require fundamentally different evaluation criteria.

    \begin{itemize}
        \item \textit{Bug fix.} For bug-fix scenarios, we extract F2P (Fail-to-Pass) and P2P (Pass-to-Pass) test cases. If a golden patch passes these tests, the data is considered valid. We then let the model act as an agent to fix the bug in a sandbox and verify correctness using both F2P and P2P tests. P2P tests are particularly important to ensure that no new bugs are introduced during the fix.
        
        \item \textit{Feature addition.} For feature additions, traditional F2P/P2P logic may not apply, since tests often depend on newly introduced code. Instead, we focus on extracting newly added test points and ensuring the golden patch passes them.
        
        \item \textit{Performance optimization.} Since performance optimization has no bug-fixing process, in such cases we extract P2P tests that can verify stable and significant performance differences before and after the optimization.
    \end{itemize}

    \item \textbf{Model-based task validation.} Raw GitHub PRs are often weakly structured, and their associated test cases may not fully specify the underlying issue, leading to ambiguous or under-specified tasks. To mitigate this, we employ a model to validate consistency between problem descriptions and test cases, and to enrich missing information when necessary, producing self-contained and executable task specifications.

    \item \textbf{Task transformations and augmentation.} To maximize dataset diversity, we apply transformation and augmentation strategies to existing PRs, generating multiple task variants from a single source.
    \begin{itemize}
        \item \textit{Bug injection.} Additional bugs are introduced into the codebase to increase task difficulty and expand the distribution of repair scenarios.
        \item \textit{Commit merging.} Adjacent commits or PRs are merged to construct multi-step repair tasks of greater complexity, following an approach similar to SWE-Smith~\citep{yang2024swesmith}.
        \item \textit{SWE-Test conversion.} Bug-fix PRs are converted into SWE-Test tasks, in which the problem formulation is inverted: rather than fixing the bug, the agent must write a test case that fails on the pre-patch code and passes after the patch is applied, directly exercising test-writing capability while remaining fully verifiable.
        \item \textit{Code review tasks.} The agent performs static analysis, inspects code changes, and identifies potential defects without requiring a runnable environment. Consistency is verified by a secondary LLM, yielding approximately verifiable tasks that contribute meaningfully to overall task diversity.
    \end{itemize}
\end{itemize}

The SWE-scaling pipeline ultimately produces a large-scale training dataset, where each instance consists of a problem statement, a test-based verifiable reward, and a runnable Docker environment. For the M2 series models, the pipeline spans more than ten programming languages and covers a wide range of coding task categories.

\subsubsection{Expert-Driven Data Collection: Application Development Tasks (AppDev)}

While SWE tasks focus on modifications to existing repositories, such as bug fixes, feature additions, and refactoring, Application Development (AppDev) tasks require building complete applications from scratch. This poses distinct challenges: tasks cannot be directly extracted from existing codebases, quality signals require runtime verification beyond static analysis, and evaluation criteria span both functional correctness and subjective design quality. To address these challenges, we design an expert-in-the-loop data pipeline that combines domain expertise with automated verification at scale. Domain experts contribute meta queries encoding production-level technical patterns, craft system prompts that guide trajectory generation, and design evaluation rubrics spanning execution, interaction, and aesthetics. An Agent-as-a-Verifier (AaaV) framework then performs automated rejection sampling by deploying generated applications in sandboxed environments and validating them against expert-defined rubrics through tool-assisted interaction. This pipeline enables us to synthesize diverse, high-quality application development trajectories across multiple domains, such as frontend, backend, mobile, desktop, and simulation.

\noindent \textbf{Expert-in-the-Loop Query Synthesis.} We synthesize diverse, high-quality development task queries through a combination of expert-designed meta queries and automated quality control. Domain experts from engineering teams contribute optimized meta queries that encode their domain knowledge. Each meta query serves as a template that captures essential technical patterns, specifying framework ecosystems (e.g., React + Zustand + Tailwind), architectural constraints, and realistic use cases grounded in production experience. Experts curate category-specific seed pools covering UI component libraries, CSS frameworks, build tools, SaaS integrations, and common application scenarios. These meta queries inject controlled variability: technology stacks, styling approaches, and functional requirements are sampled from expert-curated distributions to ensure both diversity and technical validity. The meta queries undergo iterative refinement based on downstream quality signals, allowing experts to prune patterns that consistently yield low-quality outputs and amplify those that produce well-structured development tasks.

Diverse user queries are synthesized by combining meta queries with randomly sampled seeds, and then processed through LLM generation with high temperature to maximize variation. Each meta query is expanded into multiple concrete queries that describe specific development tasks, covering feature requirements, technology choices, and UI/UX specifications. To control redundancy, we apply MinHash-based deduplication, achieving approximate near-duplicate detection in linear time with configurable thresholds. Finally, we evaluate the quality of each query by employing LLM-as-a-judge on domain-specific rubrics, organized into three categories: Tech Stack Rationality (compatibility, selection appropriateness, combination validity), Feature Feasibility (technical achievability, description specificity, UI/UX logic), and Requirement Clarity (validity, scenario authenticity, expression coherence, completeness). Queries with scores below a specific threshold are rejected. This ensures that only well-formed, technically sound, and realistically scoped development tasks proceed to trajectory sampling.

\noindent \textbf{Trajectory Sampling with Expert System Prompts.} Domain experts inject prior knowledge directly into the generation process through carefully designed system prompts. These prompts encode best practices that address known deficiencies in our early models. For example, for web frontend tasks, experts observed that the model exhibited suboptimal design habits, such as overusing template-like gradient backgrounds. To counteract these tendencies, the system prompt articulates explicit design guidelines and quality standards spanning functional completeness, code integrity, content authenticity, and aesthetics. Beyond design guidance, the prompts encourage best practices in the development process: writing specifications before implementation, maintaining structured TODO lists for complex tasks, and performing self-verification through testing. We also collect trajectories specifically targeting skill usage and internalization, enabling models to learn when and how to leverage skills effectively. A key element of our approach is prompt distillation: during trajectory sampling, the model receives the full enriched system prompt, whereas during training, we selectively drop portions of this guidance. This partial asymmetry encourages the model to internalize expert-encoded best practices as default behavior, reducing its reliance on explicit prompting at inference time.

\noindent \textbf{Rejection Sampling with Rubric-based Reward and Agent-as-a-Verifier.} Unlike SWE tasks where test cases provide natural correctness signals, application development requires holistic evaluation across multiple dimensions that cannot be assessed through static code analysis alone. We address this through Agent-as-a-Verifier (AaaV), a framework that validates trajectories by deploying the generated applications in sandboxed environments and evaluating them through tool-assisted interaction. The evaluation proceeds through three hierarchical layers, yielding binary pass/fail judgments with mandatory evidence for each criterion:

\begin{itemize}
    \item \textbf{Execution Layer} validates whether the application can actually run. Specific checks include: file existence and syntax validity, dependency resolution and installability, build success and server initialization, HTTP status and absence of JavaScript errors on page load. Trajectories failing this layer are immediately rejected, as they represent non-functional outputs.
    \item \textbf{Interaction Layer} assesses whether core functions work as intended. Using Playwright, the verifier agent checks: presence of expected interactive elements, responsiveness of buttons and forms, and end-to-end completion of core feature workflows. The agent navigates the deployed application, triggers user interactions, and observes state changes to determine whether specified functionality is actually implemented.
    \item \textbf{Visual Aesthetics Layer} evaluates subjective quality dimensions: layout professionalism, visual hierarchy clarity, color scheme harmony, and adherence to modern UI design standards.
\end{itemize}

The overall pass rate across layers serves as the reward signal for rejection sampling, with Execution Layer checks acting as hard gates for immediate rejection. This three-layer evaluation distinguishes AaaV from conventional LLM-as-a-Judge approaches: rather than assessing quality from static code or screenshots, the verifier agent actively interacts with the running application across multiple turns, scoring observed behavior against expert-defined rubrics. Crucially, the synergy between our rigorously filtered, diverse query distribution and this Agent-as-a-Verifier evaluation paradigm establishes a robust foundation for subsequent reinforcement learning, providing both a rich exploration space and reliable, environment-grounded reward signals.

\subsubsection{Terminal-Gym: Automated Task Synthesis and Environment Generation}

Beyond SWE-bench~\citep{jimenez2024swebench}, Terminal-Bench~\cite{merrill2026terminalbench} evaluates agents on realistic, complex tasks within a fully functional terminal environment, placing significantly higher demands on LLMs' system operation, debugging, and command-line capabilities. To enhance model performance on such capabilities, we propose Terminal-Gym, an automated data synthesis pipeline that systematically converts curated real-world programming scenarios into a diverse corpus of verifiable terminal tasks. By generating structured task schemas, dynamically evolving query difficulties, and automatically synthesizing robust Docker-based runtime environments, it provides a highly scalable training framework for terminal agents.

\noindent \textbf{Seed Dataset Selection.} Terminal-Gym takes the complete Stack Overflow dataset as a foundation, which provides high-quality, real-world programming and system-operation scenarios at scale. After chronologically sorting the raw data to reconstruct complete posts, we apply rigorous rule-based filtering. We discard posts lacking accepted answers, low-score posts, and overly lengthy query-answer pairs. To focus strictly on terminal scenarios, we filter by tags, retaining only threads related to terminal operations, system configuration, debugging, scripting, and relevant software engineering workflows. Each remaining post is then annotated with multiple attributes, including problem quality, task type applicability, verifiability, task category, approximate complexity, environment requirements, and execution characteristics. We select only those posts that satisfy strict criteria: they must be scriptable, terminal-compatible, verifiable, Linux/Docker-relevant, and of moderate difficulty. Finally, we discard noisy or redundant content within each thread and select a single high-quality answer, forming the foundational Query-Answer Pair.

\noindent \textbf{Query Synthesis.} We further rewrite the selected queries into structured task descriptions. This involves concretizing the execution context, including the necessary environment, required tools, expected input and output formats, and success criteria. These rewritten tasks are then graded into four tiers based on testability, completeness, and clarity; only tasks falling into the top two tiers are retained. The original Query-Answer Pairs are thus transformed into a structured schema containing a natural-language instruction, any necessary supporting files or scripts, and a succinct description of the expected terminal behavior.

\noindent \textbf{Synthesis Pipeline.} Each structured schema undergoes a three-stage transformation into a complete terminal task.

\begin{itemize}
    \item \textit{Stage 1: Environment and Test Generation.} An agent generates a Dockerfile and a corresponding test script for each task. The test is executed to verify functionality. If the test fails, structured diagnostic feedback is returned to the agent for iterative repair, which continues until the test passes or a maximum retry limit is reached.
    \item \textit{Stage 2: Query Evolution and Unified Testing.} Task instructions generated in previous steps often contain explicit hints, file paths, or expected environmental outputs. To address this, we apply a controlled query evolution process, systematically abstracting or removing these hints while ensuring semantic consistency. All resulting variants of a task are then evaluated using a unified test suite generated by LLMs. This unified testing approach forces the tests to validate the underlying logic of the task rather than overfitting to specific descriptive styles or explicit hints. Experimental results demonstrate the efficacy of this method in ensuring robust evaluation.
    \item \textit{Stage 3: Difficulty Calibration.} Finally, we rigorously filter out overly simplistic tasks. We preferentially sample task variants that contain fewer hints and exhibit lower zero-shot pass rates. This filtering process considers the historical pass rates of a reference solver and the number of repair iterations required during environment synthesis, ensuring the final benchmark remains highly challenging and discriminating.
\end{itemize}

Ultimately, Terminal-Gym provides a highly scalable training framework for complex terminal operations, a foundation we are now evolving into the zero-intervention Anything2Docker system. Furthermore, given the growing importance of code security, we are further expanding CVE-Factory~\cite{luo2026cvefactory} to encompass the broader frontier of autonomous cybersecurity research, pushing the boundaries of AI-driven vulnerability analysis and proactive defense mechanisms.

\subsection{Agentic Cowork}
\label{sec:agentic-cowork-data}

Beyond the verifiable signals available in coding and terminal tasks, real-world deployment also requires agents that can operate across heterogeneous professional environments---navigating the open web for primary sources, reasoning over financial spreadsheets, authoring presentation decks, and producing the broader range of office artifacts that end users actually consume. Agentic Cowork is the data-collection track behind these capabilities, organized around four domains: deep search and open-web research, knowledge-worker office tasks, financial analysis and spreadsheet operations, and slide generation. Although each domain operates over a distinct workspace and produces a distinct artifact, all four follow the same overall design. Tasks are instantiated on real, runnable workspaces; trajectories are distilled from a rotating set of strong teacher models under deliberately perturbed scaffolds; and acceptance is governed by a verification signal aligned with the artifact format, rather than by a single generic judge. For sub-tasks whose outcomes are not directly machine-verifiable, we collect multiple candidate responses and select among them through pairwise comparison along two axes---the reasoning-and-action trajectory and the final artifact---followed by a rubric-based filtering pass that enforces a strict accuracy and quality bar. In what follows we describe how each domain instantiates this shared pipeline.

\subsubsection{Deep Search and Open-Web Research}

This domain targets tasks that require an agent to navigate the open web, gather evidence across multiple sources, and synthesize a grounded answer. To produce such tasks at scale, we adopt a guide-and-rewrite synthesis strategy. Starting from a seed question, we iteratively rewrite the question and obscure the entities it relies on, until the task becomes difficult enough to discriminate between strong and weak agents. This procedure gives us continuous control over task difficulty, allowing easy variants to exercise basic retrieval and harder variants to demand deep, multi-step browsing and cross-source corroboration. To prevent the model from learning to fabricate plausible-sounding answers, every synthesized task is also paired with an explicit evidence specification, and a sampled trajectory is accepted only when its answer is grounded in actually retrieved evidence rather than recited from model memory. For broader, report-style queries that admit no unique short answer, we replace exact-match acceptance with a rubric-based judge that scores along factual accuracy, transparency, uncertainty handling, and risk disclosure. Trajectories are distilled from a rotating set of strong teacher models, and the surrounding scaffold is perturbed across runs so that the resulting policy generalizes beyond any single tool layout.

\subsubsection{Knowledge-Worker Office Tasks}

This domain covers the broad space of end-to-end professional deliverables---reports, slides, memos, structured documents---that knowledge workers produce as part of their daily work. We anchor the corpus to GDPval~\citep{patwardhan2025gdpvalevaluatingaimodel}, an established office-task benchmark, and extend it with a synthesis pipeline that mirrors how real professionals organize their work.

\noindent \textbf{Seed Curation.} We first curate a usable subset of canonical tasks from the seed benchmark, filtering out items that fall outside what our agent harness can support, so that the seed portion provides a clean, executable anchor for the rest of the corpus.

\noindent \textbf{Hierarchical Synthesis.} On top of this anchor, we produce a much larger self-synthesized corpus through a hierarchical, multi-stage procedure. We start from broad occupational categories sourced from public occupational databases and derive fine-grained subdivisions that incorporate cultural and regional diversity, ensuring broad coverage across industries, regions, and cultural contexts. For each subdivision, we generate concrete tasks together with detailed task descriptions that simulate real-world work scenarios, grounding the data in authentic professional activities rather than abstract or generic instructions. For each task, we further produce both a real workspace of supporting documents and several query versions at different levels of specificity, transforming high-level task descriptions into concrete, actionable problem settings and exposing the model to a spectrum of user-articulation styles. Each task additionally ships with a structured specification of the expected deliverable, so that synthesis, execution, and acceptance all share the same artifact format.

\noindent \textbf{Multi-Axis Rubric Acceptance.} Acceptance is governed by a multi-axis rubric covering positive behaviors, negative behaviors, critical errors, regional appropriateness, and depth of reasoning, applied uniformly across both the seeded and the self-synthesized portions. A typed cleanup pass further removes trajectories that fabricate data, references, or entities, so that only artifacts meeting a strict factual standard remain in the final corpus.

\subsubsection{Financial Analysis and Spreadsheet Operations}

This domain covers two complementary task families that together span the daily work of a financial professional: financial information retrieval, computation, and reasoning grounded in real financial tools; and spreadsheet operations over real workbooks. The two families are constructed by distinct task-synthesis pipelines but share the same downstream acceptance and scaffolding regime.

\noindent \textbf{Evidence-Driven Synthesis.} For the first family, we adopt the evidence-driven task-synthesis pipeline introduced in our earlier work~\citep{chen2026dive}, which inverts the conventional top-down authoring order: we first execute real financial tools to collect grounded execution traces, then reverse-derive tasks that are strictly entailed by those traces. This yields grounding by construction---every task is, by design, both executable and verifiable from observable tool outputs, and the reference answer is fully determined by what the tools actually return.

\noindent \textbf{Workbook-Walk Synthesis.} For the second family, we instead build the corpus through a workbook walk, in the spirit of trajectory-driven synthesis approaches such as~\citep{liu2025webexplorer}. An agent runs a curated set of atomic spreadsheet operations against a seed workbook, the intermediate states it traverses are recycled as new seeds, and on each resulting trajectory we synthesize tasks in reverse order, deriving the answer from the trajectory and the question from the answer. A final pass diversifies the resulting question pool along phrasing and difficulty axes.

\noindent \textbf{Coverage.} The first family targets retrieval, computation, and reasoning questions whose answers must be grounded in external financial data. The second family covers three sub-tracks: general and competition-level spreadsheet manipulation involving workbook-structure understanding, formula application, and cross-sheet operations; financial modeling spanning typical PE, VC, and M\&A scenarios; and the reconstruction of structured workbooks from semi-structured source documents.

\noindent \textbf{Acceptance.} Acceptance prefers a deterministic value-level match. Student artifacts are executed, their formulas are recalculated by an external engine, and the resulting cell values are compared against ground-truth workbooks. For deliverables whose form may legitimately vary, such as workbooks reconstructed from semi-structured documents or open-ended financial reasoning sub-tasks, we fall back to rubric-based or agent-based judging. Every task is additionally sampled under multiple scaffolds, so that the resulting policy is robust to variation in the tool interface.

\subsubsection{Slide Generation and Editing}

This domain targets both end-to-end deck creation and incremental slide editing, and the synthesis pipeline accordingly proceeds along two parallel streams. The first stream treats slide authoring as an open-ended generation problem. We curate a diverse set of source documents across business domains, and for each document derive queries that vary in description granularity, length, and language register, so that the resulting tasks span a realistic distribution of user requests. The second stream treats slide editing as a localized intervention problem. We sample real decks as seeds and generate editing instructions along multiple diversity axes, including the granularity of the edit (from individual elements through pages to the document level), the intent of the edit (content, style, or structure), and the complexity of the change. Trajectories are distilled from a rotating set of strong teacher models, with preference given to teachers whose generations exhibit the visual quality we want the student to inherit. Because slide artifacts are ultimately consumed visually, acceptance layers multiple complementary signals: execution success, functional correctness as judged by an agent, rule-based checks of basic layout aesthetics, and a final visual scorer that renders the deliverable and judges it as an image. To prevent the policy from overfitting to a single rendering toolkit, we additionally mix in trajectories produced under alternative slide-generation libraries.

\subsection{Reasoning-Intensive Tasks}
\label{sec:reasoning-data}

The core objective of reasoning data is to equip the model with the ability to engage in deep, structured thinking on complex problems---proving mathematical theorems, deriving scientific conclusions, designing algorithms, and constructing logical arguments. These tasks span numerous domains, and within every domain individual problems further admit a wide variety of valid solution strategies, resulting in a combinatorial space of tasks and approaches. This scale and diversity naturally motivates a scaling-driven approach. We scale along three complementary axes and simultaneously maintain a quality-assurance pipeline to ensure data correctness at volume.

\noindent \textbf{Query-Side Scaling.} Expanding the set of unique problems, particularly in underrepresented difficulty bands, directly improves coverage and generalization. We combine curation from existing sources with targeted synthesis of novel problems addressing skill gaps identified through error analysis.

\noindent \textbf{Response-Side Scaling.} Generating multiple correct solution paths per query improves reasoning diversity. Out-of-domain capability improves consistently as the number of responses per query increases, with benefits manifesting primarily in OOD generalization, indicating that diverse solution paths teach transferable reasoning strategies rather than solution memorization. We profile saturation characteristics across difficulty tiers and concentrate additional sampling where the model's solution diversity remains low.

\noindent \textbf{Training-Side Scaling.} Beyond expanding queries and responses independently, we study the optimal data mixture ratio---the relative proportion of query expansion versus response expansion---under fixed compute budgets. The former improves problem coverage and domain breadth, while the latter deepens the model's command of the solution strategy space. We empirically calibrate this mixture per training stage based on where the current capability bottleneck lies, and adopt dynamic allocation that concentrates resources on the model's weak areas.

\noindent \textbf{Quality Assurance.} Scaling at volume introduces the risk of noise and incorrect data. To maintain correctness, we enforce quality control across every stage of the pipeline. For \textit{queries}, we apply multi-stage cleaning that combines direct query tagging with cross-comparison of rollout responses to identify ambiguous or ill-formed problems. For \textit{verifiers}, we conduct systematic case analysis to cover more boundary conditions and edge cases, ensuring verification logic remains accurate across diverse problem types. For \textit{answers}, we cross-check correctness by comparing performance differentials across multiple models, flagging instances where disagreements indicate potential labeling errors. For \textit{responses}, we apply a structured rubric-based scoring framework that evaluates reasoning traces along well-defined quality dimensions before inclusion in the training corpus.

\subsection{General-Purpose Conversation and Writing}

This track complements the agentic and reasoning corpora above with broad-coverage conversational data spanning writing, general question answering, and multi-turn dialogue. The corpus primarily consists of high-quality samples featuring long chain-of-thought (long CoT) reasoning, aimed at instilling reasoning capability into the model while preserving its general-purpose competence and providing a stable cold-start foundation for subsequent reinforcement learning.

Each sub-domain carries a distinct emphasis. For \textit{writing}, the primary focus is on style: high-quality queries are carefully curated so that responses adhere to a specific stylistic standard, capturing the nuances of tone, structure, and expression. A file system is additionally incorporated into the writing pipeline, enabling the model to read from and write to structured documents and thereby aligning its writing capabilities with real-world productivity scenarios. For \textit{general question answering}, where queries tend to be relatively straightforward, the focus shifts to selecting among multiple candidate responses to satisfy preference-aligned quality requirements. For \textit{multi-turn dialogue}, the emphasis lies in instruction- and rubric-following under more complex interactive settings---sustained coherence across many turns, cross-turn context tracking, and robust comprehension over long contexts.

To further enhance the model's capability and robustness, the dataset includes both tool-augmented and tool-free samples. Tool-augmented samples teach the model to effectively leverage external tools such as code interpreters and search engines when appropriate, while tool-free samples ensure that the model can reason independently without relying on external assistance. This balanced composition enables the model to adapt flexibly across diverse interaction scenarios.

After generation, all data undergoes rigorous verification through automated verifiers (rule-based checkers and model-based evaluators), along with systematic quality checks to ensure both correctness and consistently high quality before being incorporated into the training pipeline.

\subsection{Role-Play and Persona Coherence}

To support persona-conditioned long-horizon dialogue---a major real-world deployment mode that the preceding general-conversation track does not exercise---we treat role-play as a distinct data track with its own formalization, benchmark, synthesis pipeline, and reward signal.

We formalize role-play~\citep{minimax2026deepdive} as long-horizon conditional generation over the joint space $\{\text{Worlds}\} \times \{\text{Stories}\}$, conditioned on $\{\text{User Preferences}\}$. The core objective is to maintain physical, narrative, and stylistic coherence across extended multi-turn conversations. Based on the insight that misalignment is objectively detectable while alignment is subjective, we introduce \textit{Role-Play Bench}, which evaluates multi-turn self-play trajectories by penalizing specific failure modes (e.g., out-of-character breaks, logic errors), yielding offline metrics that strongly correlate with online engagement. We synthesize training data via large-scale self-play between stylistically diverse expert models. To ensure quality and prevent mode collapse, we apply dispersion sampling across four axes, Best-of-N filtering, and periodic segment-level rewriting by an LLM-as-a-judge. We optimize the policy via RLHF using implicit and explicit feedback from real product interactions; these raw signals are denoised through causal inference and stratified bias removal to isolate genuine quality indicators, with entropy monitoring applied to mitigate reward hacking.

%% file: section/post_training_sft.tex
\section{Supervised Fine-Tuning}

We conduct Supervised Fine-Tuning (SFT) to instill the desired interleaved thinking behavior in M2, providing a strong starting point for the subsequent RL stage. Unlike conventional long Chain-of-Thought (CoT) data where reasoning is confined to a single contiguous block, our SFT data interleaves thinking traces with intermediate actions and observations, enabling the model to reason, act, and revise within a unified trajectory. To produce such data at scale, we build a systematic data pipeline that performs large-scale rejection sampling against domain-specific reward signals followed by multi-stage data cleaning, yielding high-quality interleaved-thinking trajectories. The resulting SFT corpus spans four core domains---chat, reasoning, code, and cowork---covering both single-turn reasoning and multi-turn agentic interactions.

%% file: section/post_training_rl.tex
\section{Reinforcement Learning}
\label{sec:rl}

\subsection{RL Algorithm}

\subsubsection{Agent RL Modeling}

We formulate agent reinforcement learning by treating the LLM as a policy and everything outside the model's generation process---including context management, memory access, and agent state transition---as the environment. This separation provides a clean abstraction that naturally extends the standard RL framework to accommodate the complexity of agentic systems.

\subsubsection{MDP Formulation}

We model the agent-environment interaction as a Markov Decision Process $M = (S, A, T, R, \gamma)$. At each step $t$, the agent observes a state $s_t \in S$---comprising the current context window content, including the task instruction, prior conversation history, tool outputs and any artifacts produced during the agent loop---and produces an action $a_t \in A$, defined as a single-step LLM completion. This completion may contain natural language reasoning, a tool invocation request, an explicit context management operation, a communication with a sub-agent, or any combination thereof. The environment then executes the requested operations and returns an observation $o_t$, which, together with possible context management operations, determines the next state:
\begin{equation}
s_{t+1} = f_{\text{trans}}(s_t, a_t, o_t),
\end{equation}
where $f_{\text{trans}}$ denotes an arbitrary state transition function that may change the accumulated context and the internal state of the agent loop. The trajectory $\tau = (s_0, a_0, s_1, a_1, \ldots, s_T, a_T)$ constitutes a complete episode, and the policy $\pi_\theta(a_t \mid s_t)$ is parameterized by the LLM weights $\theta$.

A key design principle is that the environment boundary is drawn at the model's generation interface. All components that process, transform, or respond to the model's outputs are treated as part of the environment dynamics:
\begin{itemize}
    \item \textbf{Tool Environments:} external tool execution (code interpreters, search engines, APIs) that returns structured observations in response to tool-call actions.
    \item \textbf{Agent Harness:} the harness-level control flow that governs how the agent proceeds between LLM calls---including context management, branching logic, sub-agent delegation, and interactions with external modules.
\end{itemize}

\subsubsection{Training Objective}

A key consequence of this modeling is that $\pi_\theta$ is not required to explicitly reason about or control the environment and state transitions. Training operates on individual $(s_t, a_t)$ pairs as atomic units: each pair constitutes a single training sample for the policy gradient. This decouples the policy from the mechanics of state evolution---the model need not be aware of whether $s_t$ resulted from a simple message append, an aggressive context truncation, or a complete history rewrite. Meanwhile, credit assignment, advantage estimation, and reward propagation can still be performed at the episode level over the full trajectory $\tau$, ensuring that the contribution of each $(s_t, a_t)$ pair is evaluated in the context of the overall task outcome.

\subsubsection{Policy Optimization}

\noindent \textbf{CISPO.} We adapt Clipped Importance Sampling Policy Optimization (CISPO)~\citep{minimax2025m1} to M2 series RL training. The objective function is:
\begin{equation}
J_{\text{CISPO}}(\theta) = \mathbb{E}_{(q,a) \sim \mathcal{D},\, \{o_i\}_{i=1}^G \sim \pi_{\theta_{\text{old}}}(\cdot \mid q)} \left[ \frac{1}{\sum_{i=1}^G |o_i|} \sum_{i=1}^G \sum_{t=1}^{|o_i|} \mathrm{sg}\!\left(\hat{r}_{i,t}(\theta)\right) \hat{A}_{i,t} \log \pi_\theta(o_{i,t} \mid q, o_{i,<t}) \right],
\end{equation}
where $G$ is the number of rollout trajectories per prompt, $|o_i|$ is the token length of trajectory $i$, and $\mathrm{sg}(\cdot)$ denotes the stop-gradient operator that prevents gradient flow through the importance weight.

The importance sampling ratio is clipped asymmetrically:
\begin{equation}
\hat{r}_{i,t}(\theta) = \mathrm{clip}\!\left(\frac{\pi_\theta(o_{i,t} \mid q, o_{i,<t})}{\pi_{\theta_{\text{old}}}(o_{i,t} \mid q, o_{i,<t})},\; 0,\; 1 + \epsilon_{\text{high}}^{\text{IS}}\right).
\end{equation}
The upper bound $1 + \epsilon_{\text{high}}^{\text{IS}}$ prevents excessively large policy updates, while the zero lower bound permits aggressive down-weighting of actions that become improbable under the current policy. The stop-gradient on the clipped ratio ensures that the importance weight modulates the gradient magnitude without introducing second-order terms, yielding a stable first-order update rule.

The advantage estimate is computed via reward-to-go with a trajectory-level baseline:
\begin{equation}
\hat{A}_{i,t} = \sum_{p=t}^{T} r_p - B_i,
\end{equation}
where $r_p$ is the composite reward at step $p$ (defined below) and $B_i$ is the baseline computed over trajectory $i$ for variance reduction.

\subsubsection{Reward Design}

Standard outcome-based rewards are insufficient for credit assignment in agent trajectories that may span up to 192K tokens with thousands of intermediate actions. We design a composite reward framework with three components.

\noindent \textbf{Process Reward.} We assign dense, intermediate rewards that target specific behavioral patterns throughout the trajectory, including penalties for language mixing and tool invocation format errors, and rewards for well-structured intermediate reasoning steps. These process rewards provide fine-grained supervisory signal at each $(s_t, a_t)$ pair, substantially improving credit assignment granularity over sparse outcome-only feedback.

\noindent \textbf{Task Completion Time Reward.} Traditional RL objectives optimize solely for correctness, neglecting execution efficiency. For agentic tasks, functionally equivalent trajectories may differ dramatically in wall-clock latency due to sequential versus parallel tool execution and sub-agent invocation overhead. We incorporate relative completion time as an explicit reward:
\begin{equation}
r^{\text{speed}}_t = h\!\left(\frac{T_{\text{completion}}}{T_{\text{baseline}}}\right),
\end{equation}
where $h(\cdot)$ is a monotonically decreasing shaping function, $T_{\text{completion}}$ is the wall-clock time taken by the rollout, and $T_{\text{baseline}}$ is a reference completion time. This incentivizes the policy to discover and exploit parallelism opportunities, producing solutions that are both correct and efficient.

\noindent \textbf{Reward-to-Go with Baseline.} To reduce gradient variance in long-horizon tasks, we adopt a reward-to-go formulation:
\begin{equation}
G_t = \sum_{\tau=t}^{T} \gamma^{\tau-t} r_\tau.
\end{equation}
Combined with the trajectory-level baseline, this formulation concentrates gradient signal on actions whose consequences are not yet accounted for, improving credit assignment precision and stabilizing the optimization.

The composite reward at each step is:
\begin{equation}
r_t = \alpha \cdot r_t^{\text{process}} + \beta \cdot r_t^{\text{speed}} + r_t^{\text{perf}},
\end{equation}
where $\alpha$ and $\beta$ are coefficients balancing dense behavioral feedback and efficiency incentives against the primary task performance signal.

\subsubsection{Mixed-Domain RL Training}

A critical challenge in training general-purpose agents is avoiding the trade-off between task-specific optimization and broad capability preservation. Single-domain RL training---fine-tuning exclusively on agentic tasks---risks catastrophic forgetting of the model's foundational reasoning and general knowledge capabilities. Conversely, sequential multi-stage training across domains induces negative transfer, as gains in one domain erode performance in previously trained domains.

We adopt a mixed-domain RL training strategy that addresses both issues. Training proceeds through multiple stages, and within each stage, training data is drawn simultaneously from four domains: reasoning, coding, agent, and general. This joint optimization ensures that the policy gradient updates are informed by a diverse task distribution at every training step, preventing the optimizer from overfitting to any single domain's reward landscape.

Across stages, we systematically adjust three axes:
\begin{itemize}
    \item \textbf{Domain mixing ratios.} The relative proportions of data from each domain are tuned per stage. Early stages emphasize foundational capabilities (reasoning and general domains) to consolidate the model's base competence, while later stages progressively increase the proportion of agent and coding tasks to sharpen task-specific performance.
    \item \textbf{Context length.} We expand the maximum context length at a per-domain granularity across stages. This curriculum-style progression enables the model to first master short-horizon decision-making before extending to the long-context trajectories characteristic of complex agent tasks.
    \item \textbf{Difficulty distribution.} Within each domain, the difficulty distribution of training tasks shifts progressively toward harder instances. Early stages include a broad mix to establish robust foundations, while later stages concentrate on challenging scenarios that push the policy's frontier.
\end{itemize}

This mixed-domain strategy yields compounding benefits: it simultaneously improves the model's foundational reasoning ability, task-specific quality across all target domains, and end-to-end user experience---since agents deployed in practice encounter a heterogeneous mix of requests that spans all four domains.

\subsection{RL Infrastructure}

\subsubsection{Problem Formulation}

Training RL agents at scale requires simultaneously satisfying three desiderata that are fundamentally in tension---the ``impossible triangle'':
\begin{itemize}
    \item \textbf{System Throughput:} maximizing the raw tokens processed per unit time, governed by rollout latency, training iteration time, data processing overhead, and I/O bandwidth.
    \item \textbf{Training Stability:} bounding the variance of policy gradient updates to ensure monotonic improvement and convergence, i.e., $\mathbb{E}[\mathrm{Var}(\nabla_\theta J)] < \delta$.
    \item \textbf{Agent Flexibility:} supporting arbitrary agent architectures $\mathcal{A} \in \Omega_{\text{agent}}$---from simple single-turn scaffolds to complex multi-agent systems with dynamic context management---without requiring agent-specific modifications to the training framework.
\end{itemize}

We formulate the infrastructure optimization objective as maximizing the Effective Agent Training Yield:
\begin{align}
\max_\theta\ &J(\theta) = \mathrm{Throughput}(\mathcal{A}) \times \mathrm{SampleEfficiency}(\mathcal{A}), \\
\text{s.t.}\quad &\forall \mathcal{A} \in \Omega_{\text{agent}},\quad \mathbb{E}[\mathrm{Var}(\nabla_\theta J)] < \delta,\quad \mathbb{E}[\|J^{(T)} - J^*\|] < \epsilon. \nonumber
\end{align}

Each pair of these three goals creates a specific engineering tension. Maximizing throughput under heterogeneous agent rollout times (ranging from seconds to hours) conflicts with maintaining distributional consistency for stable training. Supporting arbitrary agent architectures conflicts with the tight coupling between agent state and training logic required for efficient data processing. Achieving stable credit assignment in long-horizon trajectories (up to 192K tokens) conflicts with throughput-optimal scheduling that biases toward short, easy tasks. The Forge infrastructure resolves these tensions through the architectural and algorithmic designs described below.

\subsubsection{System Architecture}

\begin{figure}[h]
\centering
\includegraphics[width=0.78\textwidth]{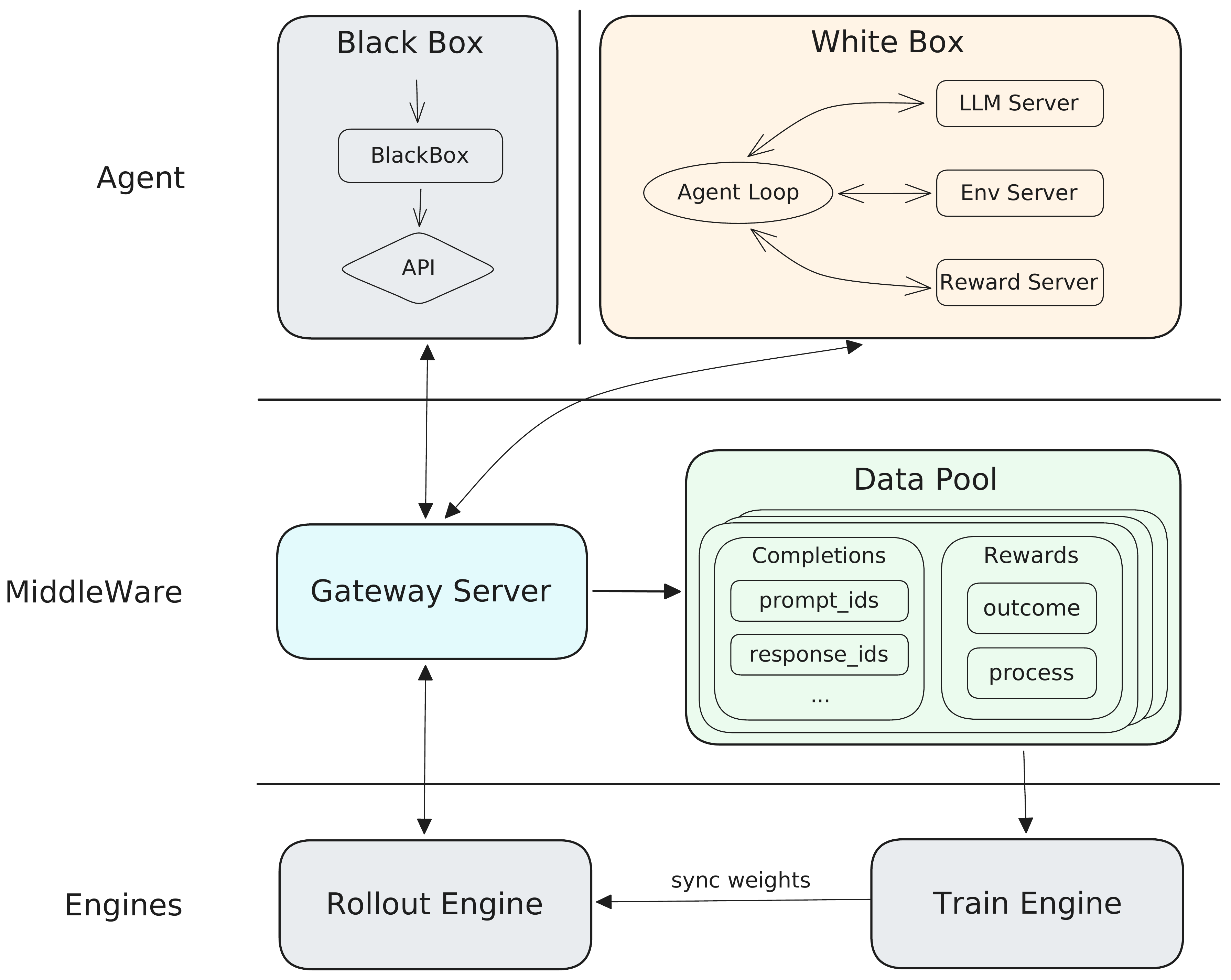}
\caption{Overview of the Forge RL system. Three decoupled modules---Agent Side, Training/Inference Side, and the middleware abstraction layer (Gateway Server, Data Pool)---communicate through standardized interfaces, allowing the agent and the training loop to scale independently.}
\label{fig:forge-arch}
\end{figure}

The Forge system is organized into three decoupled modules connected through a middleware abstraction layer (Figure~\ref{fig:forge-arch}). This architecture directly operationalizes the agent RL modeling described in the preceding subsection: the boundary between policy and environment---drawn at the model's generation interface---maps onto the boundary between the Training/Inference Side and the Agent Side.

\noindent \textbf{Agent Side.} This module encapsulates arbitrary agent implementations and orchestrates their interactions with external environments. Functioning as a pure trajectory producer, the agent side is entirely agnostic to training and inference mechanics. It drives the environment dynamics defined in the MDP formulation---executing tool calls, managing context, and accessing memory---and records the resulting $(s_t, a_t, o_t)$ tuples for downstream training consumption.

\noindent \textbf{Middleware Abstraction Layer.} Two components bridge the agent and training sides. The Gateway Server provides a standardized communication interface that routes completion requests between agents and the LLM engine, abstracting away the heterogeneity of agent scaffolds. The Data Pool serves as distributed trajectory storage that asynchronously collects rollout data, fully decoupling the generation and training pipelines and enabling each to scale independently.

\noindent \textbf{Training and Inference Side.} The Rollout Engine handles high-throughput token generation, serving as the policy $\pi_\theta$ that responds to Gateway requests. The Train Engine consumes processed trajectory sequences from the Data Pool, computes the CISPO policy gradient, and synchronizes updated weights back to the Rollout Engine.

\subsubsection{White-Box and Black-Box Agent Support}

The MDP formulation in the preceding subsection defines context management, tool execution, and memory access as environment dynamics. The infrastructure must support two distinct paradigms for how agents implement these dynamics, which differ in the degree of visibility the framework has into the agent's internal state.

\noindent \textbf{White-Box Agents} expose their context management logic to the training framework. The state transition $s_{t+1} = f_{\text{CM}}(\mathrm{concat}(s_t, a_t, o_t))$ is implemented within the framework itself, enabling the training pipeline to directly observe and backpropagate through context transformation operations. This tight integration allows the framework to construct training sequences that faithfully reflect the CM-induced distribution, ensuring that the policy is optimized over the true inference-time state distribution. White-box integration is particularly effective for agents with well-defined CM strategies (e.g., sliding-window truncation, periodic summarization), where the framework can reconstruct exact training states.

\noindent \textbf{Black-Box Agents} treat the agent as an opaque trajectory producer. Agents route completion requests to the RL service Gateway without exposing their internal context management, memory compression, or multi-agent coordination logic. The framework collects only the externally visible $(s_t, a_t, o_t)$ tuples---where $s_t$ is the context as presented to the LLM at each completion request---and constructs training data from these observations. This non-intrusive paradigm supports arbitrary internal architectures, including deep thinking loops, aggressive context rewriting, and hierarchical multi-agent systems, delivering consistent improvements without requiring any agent-side modifications.

The Gateway-based abstraction unifies both paradigms: white-box agents differ only in that their CM operations are registered with the framework for training-time reconstruction, while black-box agents rely solely on the observed request stream. This design has been validated across hundreds of distinct agent scaffolds and thousands of tool invocation formats.

\subsubsection{Windowed FIFO Scheduling}

Agent rollout completion times exhibit extreme variance---from seconds for simple API calls to hours for complex reasoning chains---creating a fundamental tension between throughput and distributional consistency. Strict FIFO scheduling preserves the data distribution but suffers from straggler-induced head-of-line (HoL) blocking. Fully greedy scheduling (fetching whichever trajectory completes first) maximizes throughput but causes severe distribution shift: early training batches are dominated by short, easy tasks, while hard tasks cluster in later batches, leading to gradient oscillation and optimization instability.

\begin{figure}[h]
\centering
\includegraphics[width=0.95\textwidth]{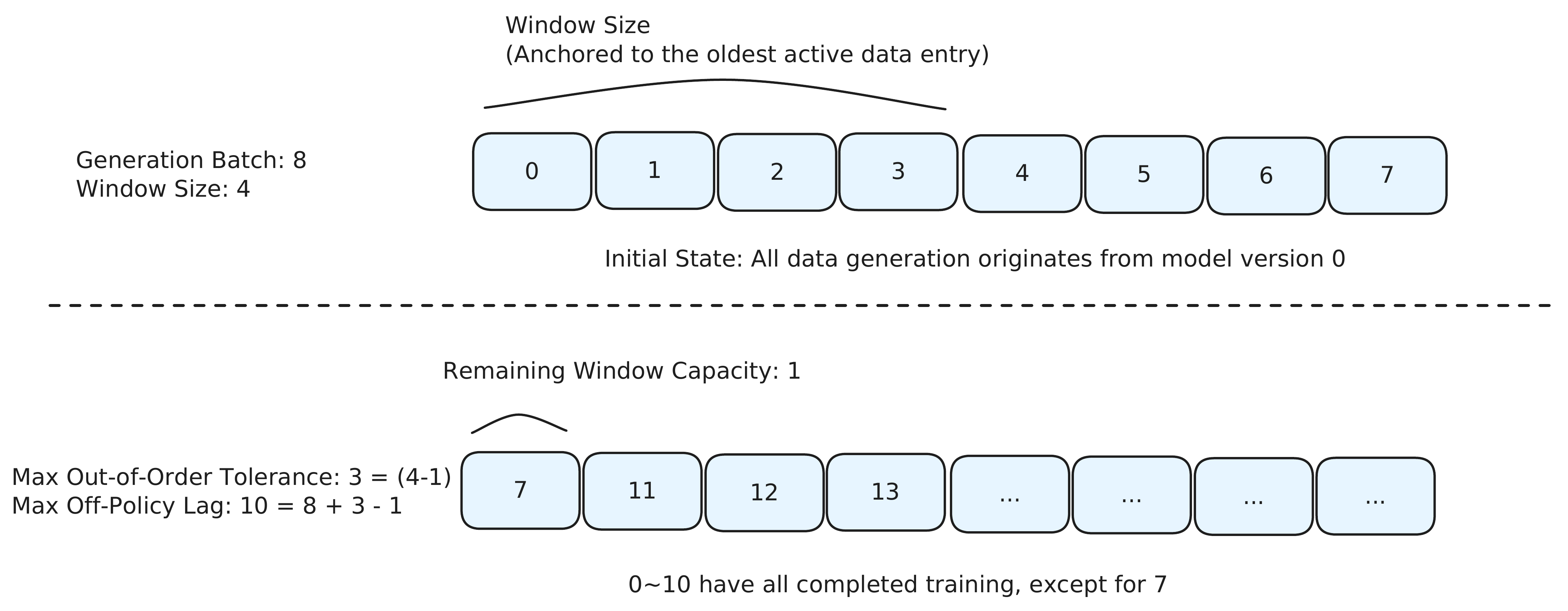}
\caption{Windowed FIFO scheduling. The training scheduler only fetches completed trajectories within a sliding window of size $W$ over the generation queue: within the window, completion order is free (mitigating head-of-line blocking); across window boundaries, strict FIFO is enforced (preserving distributional consistency).}
\label{fig:window-fifo}
\end{figure}

We propose \textbf{Windowed FIFO} (Figure~\ref{fig:window-fifo}), a hybrid scheduling strategy that interpolates between these extremes. Given a generation queue $Q = [T_0, T_1, \ldots, T_{N-1}]$ with current head index $i$, the training scheduler may only fetch completed trajectories within a sliding window $[T_i, T_{i+W-1}]$, where $W$ is the window size (e.g., $W = 0.3N$). Within the window, the scheduler operates greedily---any completed trajectory may be fetched immediately, mitigating HoL blocking. Across window boundaries, strict ordering is enforced: trajectories beyond the window are blocked regardless of completion status, and the window advances only as head-of-window tasks are consumed.

This provides a tunable trade-off controlled by $W$: smaller values approach strict FIFO (maximum distributional consistency), while larger values approach greedy scheduling (maximum throughput). In practice, $W = 0.3N$ maintains near-FIFO distributional properties while substantially reducing cluster idle time.

\subsubsection{Prefix Tree Merging for Training Acceleration}

In multi-turn agent trajectories, sequential message appending and context management operations produce extensive shared prefixes across training samples within the same rollout group. Traditional training treats each sample independently, redundantly recomputing common prefixes---a particularly severe source of waste in long-context agent scenarios.

\begin{figure}[h]
\centering
\includegraphics[width=0.95\textwidth]{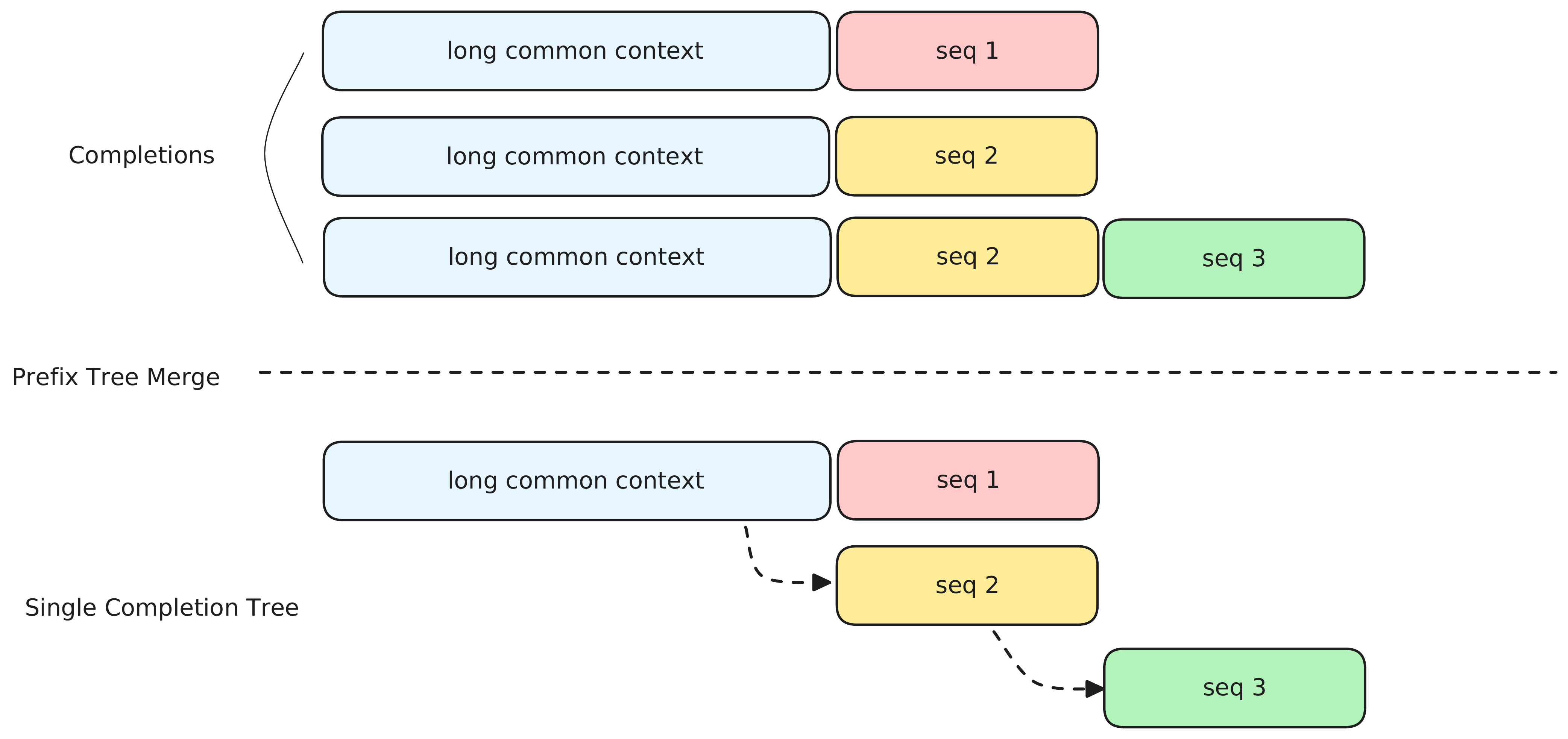}
\caption{Prefix tree merging. Completions that share a common prefix within a training batch are merged into a tree; the shared prefix is computed exactly once in the forward pass and the computation branches into individual response segments. The tree is deconstructed before the loss is computed independently per sample, so the procedure is mathematically equivalent to independent-sample training while eliminating redundant prefix recomputation.}
\label{fig:prefix-tree-merge}
\end{figure}

We propose \textbf{prefix tree merging} (Figure~\ref{fig:prefix-tree-merge}), which restructures training computation from linear sample processing to tree-structured computation. Multiple completions sharing a common prefix are merged into a single prefix tree: the shared prefix is computed exactly once in the forward pass, after which the computation branches into individual response segments. After the forward pass, the tree is deconstructed using stored metadata, and loss is computed independently per sample.

This restructuring is mathematically equivalent to independent-sample training---the causal attention computation over a shared prefix produces identical activations regardless of whether it is computed once or $k$ times---guaranteeing zero approximation error. In practice, prefix tree merging achieves up to $40\times$ training speedup with corresponding reductions in memory consumption, enabling longer sequences and larger batch sizes without any compromise in training fidelity.

\subsubsection{Inference Acceleration}

We employ three architectural optimizations to maximize generation throughput for agent RL rollouts.

\noindent \textbf{MTP-based Speculative Decoding.} Multi-Token Prediction (MTP) modules are continuously co-trained with the RL policy via top-$K$ KL divergence loss. This co-training ensures that draft acceptance rates remain high throughout the non-stationary RL optimization process, preventing the distribution shift that would otherwise degrade speculative decoding performance as the policy evolves.

\noindent \textbf{Heterogeneous Prefill-Decode Disaggregation.} Prefill and decode operations are decoupled into independently scheduled instances, eliminating the mutual interference that arises from mixed scheduling in Mixture-of-Experts (MoE) architectures. Each phase adopts parallelism strategies optimized for its computational profile, simultaneously maximizing global throughput and minimizing tail latency.

\noindent \textbf{Global L3 KV Cache Pool.} A distributed, DFS-backed global KV cache maximizes prefix cache hit rates through group-level rollout scheduling. A cost-aware request router dynamically balances queuing delay against cache migration costs, maximizing cache locality without overloading individual instances and avoiding redundant prefilling across the multi-turn interactions characteristic of agent RL.

%% file: section/post_training_agentic.tex
\section{Agentic Mechanism}

\subsection{Interleaved Thinking}
\label{sec:interleaved-thinking}

LLM agents must orchestrate multi-step workflows that alternate between natural-language reasoning and external tool invocations such as code execution, web browsing, and API calls. A critical design question is how the model's chain-of-thought (CoT) should interact with tool-use turns, and whether prior reasoning state should be preserved across interaction rounds. In MiniMax-M2, we adopt \textbf{interleaved thinking} as a first-class agent modeling principle: the model alternates between explicit deliberation and tool execution within a single trajectory, carrying the full reasoning state forward across turns.

\noindent \textbf{Interleaved chain-of-thought.} We define interleaved thinking as a generation protocol in which the model produces reasoning tokens $r_t$ and action tokens $a_t$ in an alternating sequence:
\begin{equation}
\tau = (r_1, a_1, o_1, r_2, a_2, o_2, \ldots, r_T, a_T, o_T),
\end{equation}
where $o_t$ denotes the observation (tool output) returned after executing action $a_t$. Each reasoning segment $r_t$ is conditioned on the full history $(r_1, a_1, o_1, \ldots, r_{t-1}, a_{t-1}, o_{t-1})$, allowing the model to revise plans, update hypotheses, and incorporate new evidence before selecting the next action. This contrasts with two common alternatives: (1) front-loaded reasoning, where all reasoning tokens are produced before any actions, preventing adaptation to intermediate observations; and (2) stateless per-turn reasoning, where prior reasoning tokens $r_{<t}$ are stripped from context before generating $r_t$, preventing the model from building on earlier analysis.

\noindent \textbf{Reasoning state persistence.} The key architectural decision enabling interleaved thinking is that the complete model output from turn $t$---including all thinking blocks---is appended to the message history and provided as context for turn $t+1$. Let $\mathcal{H}_t$ denote the conversation history at turn $t$. Under reasoning state persistence:
\begin{equation}
\mathcal{H}_{t+1} = \mathcal{H}_t \oplus [\mathrm{assistant}(r_t, a_t)] \oplus [\mathrm{tool}(o_t)].
\end{equation}
When reasoning state is dropped, history degrades to $\mathcal{H}_{t+1}^{\text{(drop)}} = \mathcal{H}_t \oplus [\mathrm{assistant}(a_t)] \oplus [\mathrm{tool}(o_t)]$, forcing the model to re-derive context, constraints, and partial conclusions at every turn, leading to cumulative state drift and degraded self-correction.

\begin{figure}[h]
\centering
\includegraphics[width=0.65\textwidth]{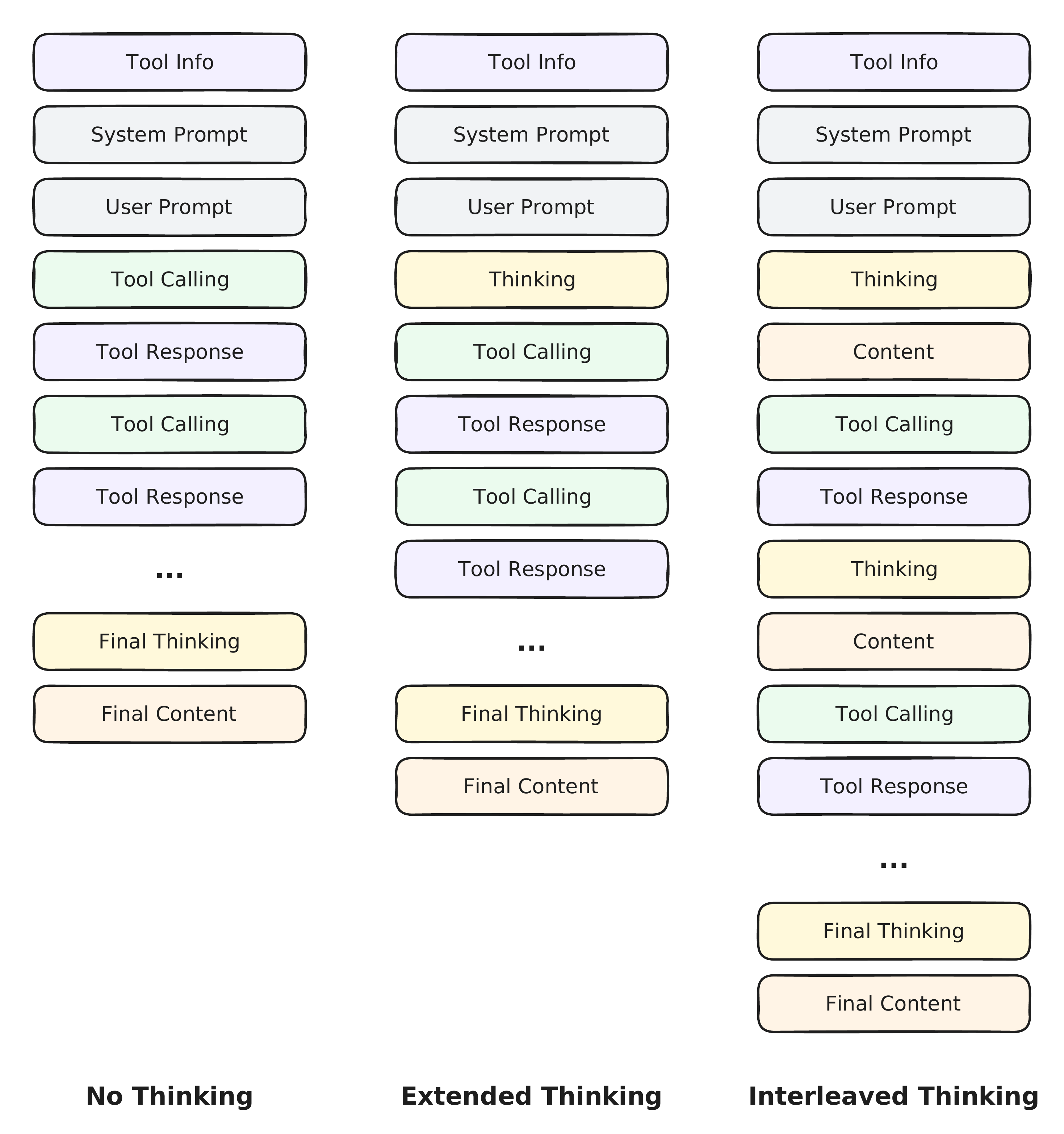}
\caption{The Plan-Act-Reflect loop of interleaved thinking in M2.}
\label{fig:interleaved-thinking}
\end{figure}

\noindent \textbf{The Plan-Act-Reflect loop.} Interleaved thinking operationalizes a structured cognitive loop at each turn $t$: (1) \textbf{Plan}---the model reviews accumulated state from prior reasoning and observations, then formulates or refines a strategy; (2) \textbf{Act}---the model selects and executes a tool call grounded in the plan; (3) \textbf{Reflect}---the model evaluates the observation against expectations, updates its world model, and determines whether to revise the plan or proceed. This loop enables self-correction through reflection on unexpected observations, sample efficiency by reusing hypotheses and intermediate conclusions rather than re-deriving them, and debuggability via the interpretable reasoning trace. Figure~\ref{fig:interleaved-thinking} illustrates this process.

\noindent \textbf{Effect of reasoning state persistence.} We ablate reasoning state persistence by stripping thinking blocks from prior turns before each model invocation, yielding consistent gains across agentic benchmarks, with particularly large improvements on tasks requiring extended multi-step reasoning like deep search and software engineering, indicating that interleaved thinking is most impactful when tasks demand sustained planning and iterative refinement across many steps.

\subsection{Self-Evolution}
\label{sec:self-evolution}

\begin{figure}[h]
\centering
\includegraphics[width=0.9\textwidth]{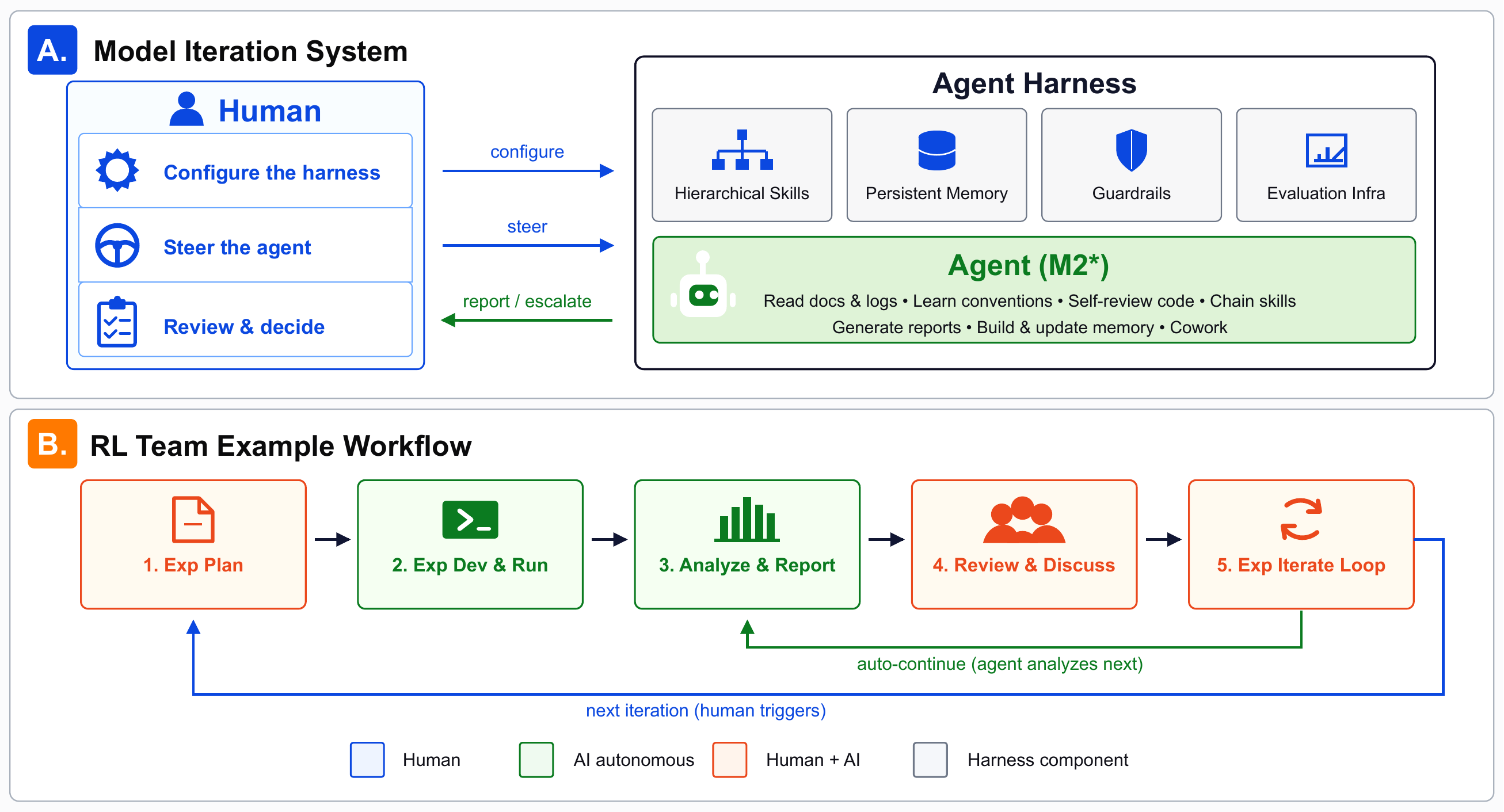}
\caption{(A) The Model Iteration System used to drive M2.7's autonomous execution. (B) The dual-loop workflow used by our RL team.}
\label{fig:self-evolution}
\end{figure}

\noindent We detail a shift in model training toward self-evolution. Rather than a sudden leap, this shift represents the culmination of the M2 series' steadily growing agentic capabilities---progressing from routine debugging and reporting tasks into a fully integrated pipeline where M2.7 actively drives its own iterative development (Figure~\ref{fig:self-evolution}).

\noindent To operationalize this, we developed the \textbf{Model Iteration System} (Figure~\ref{fig:self-evolution}A), which operates on the principle that humans steer while models build. Researchers configure goals, guide the agent via chat, and review outputs to decide the next steps. The agent functions within an ``Agent Harness''---a workspace generated entirely by an internal M2.7 model with zero human-written code. This harness equips the model with hierarchical skills for action chaining, persistent memory, safety guardrails, and evaluation infrastructure.

\noindent In practice, our RL team uses this system through a dynamic, dual-loop workflow (Figure~\ref{fig:self-evolution}B). Following human-led experiment planning, M2.7 enters an autonomous execution phase to profile ongoing runs, read logs, and diagnose metric anomalies. By automatically debugging code and adjusting configurations, the model directly intervenes in its training loop, absorbing 30\% to 50\% of the daily iteration workload. Human review triggers major iteration decisions, while the agent can auto-continue bounded analysis between reviews.

\noindent Ultimately, this system enables recursive scaffold upgrades. Tasked with optimizing an internal programming scaffold, M2.7 executed a fully autonomous 100-round iteration cycle: analyzing failures, modifying code, and evaluating changes. This exploration introduced mechanisms such as loop detection and discovered better parameter combinations, yielding a 30\% performance gain on in-house evaluations and showing that the model can improve the infrastructure shaping its subsequent iterations.

%% file: eval.tex
\section{Evaluation}
\label{sec:eval}

We evaluate MiniMax-M2.7 against the strongest publicly deployed closed-weight frontier reasoning models---Claude Opus 4.6, Claude Sonnet 4.6~\citep{anthropic2025claude46}, GPT 5.4~\citep{openai2025gpt5}, and Gemini 3.1 Pro~\citep{google2025gemini31}---across three capability areas that mirror the design priorities of the M2 series: agentic coding, agentic cowork (search, multi-tool agents, and workspace operation), and reasoning \& knowledge. To make the within-series progression explicit, we additionally report scores for the previous public release, MiniMax-M2.5. Throughout, we deliberately favor benchmarks that exercise long-horizon, environment-grounded behavior over closed-form QA, since those are the regimes the M2 data pipelines and Forge RL system are explicitly built for.

\subsection{Evaluation Settings}

\noindent \textbf{Models and inference modes.} Each closed-weight baseline is evaluated in its strongest reasoning configuration: Claude Opus 4.6 and Claude Sonnet 4.6 with extended thinking, GPT 5.4 with high reasoning effort, and Gemini 3.1 Pro with high reasoning effort. M2.7 and M2.5 are evaluated with thinking enabled and the interleaved-thinking trajectory protocol described in \S\ref{sec:interleaved-thinking}. Unless otherwise noted, generation uses temperature 1.0 and top-p 0.95; on agentic benchmarks all models share the same scaffold and tool environment.

\noindent \textbf{Benchmarks.} We organize the reported benchmarks into five blocks:

\begin{itemize}
    \item \textbf{Software engineering and coding agent.} \emph{SWE-bench Pro}~\citep{swebenchpro2025} for industry-grade repository repair; \emph{SWE-bench Multilingual}~\citep{rashid2025swebenchmultilingual} for cross-language repository-level editing; \emph{Multi-SWE-bench}~\citep{multiswebench2025} for multi-repo task transfer; \emph{NL2Repo}, our internal natural-language-to-repository synthesis benchmark; \emph{Terminal-Bench 2.0}~\citep{merrill2026terminalbench} for terminal and system-operation tasks; and \emph{MLE Bench Lite}~\citep{chan2025mlebench} for autonomous machine-learning engineering, which we revisit as a self-evolution case study in \S\ref{sec:eval-case}.

    \item \textbf{Application development.} \emph{VIBE-Pro}, our internal end-to-end full-stack application development benchmark; and \emph{HyperTask}, an internal long-horizon ``vibe coding'' suite of $\sim$100 feature- or step-level requirements per task.

    \item \textbf{Cowork --- search and deep research.} \emph{BrowseComp}~\citep{wei2025browsecomp}, \emph{Wide Search}~\citep{liu2025widesearch}, and our internal \emph{RISE} benchmark for multi-step browsing under realistic web complexity.

    \item \textbf{Cowork --- agent, office, and workspace.} \emph{GDPval-AA}~\citep{patwardhan2025gdpvalevaluatingaimodel}, the Artificial-Analysis-judged subset of OpenAI's GDPval office-task suite; \emph{Toolathlon}~\citep{huang2025toolathlon} for heterogeneous tool use; \emph{MM Claw}, our internal multi-modal office-claw benchmark; \emph{MEWC v2}, an internal hard-task subset of 100 problems drawn from the Microsoft Excel World Championship pool; and \emph{Finance Modeling Pro}, an internal Excel-grounded financial-modeling suite scored by expert rubrics.

    \item \textbf{Reasoning and knowledge.} \emph{AIME 2026}~\citep{aime2026} for competition mathematics; \emph{GPQA-Diamond}~\citep{rein2024gpqa} for graduate-level science; \emph{SciCode}~\citep{tian2024scicode} for scientific code generation; \emph{IFBench}~\citep{pyatkin2025ifbench} for instruction following; \emph{AA-LCR} (Artificial Analysis Long-Context Reasoning) for retrieval-and-reasoning over long inputs; \emph{HLE}~\citep{phan2025humanity} (Humanity's Last Exam, no-tool subset) for frontier-difficulty open knowledge; and \emph{MMLU-Pro}~\citep{wang2024mmlu} for broad knowledge.
\end{itemize}

\noindent \textbf{Evaluation configurations.} \emph{SWE-bench Pro}, \emph{SWE-bench Multilingual}, \emph{Multi-SWE-bench}, and \emph{NL2Repo} are run on internal infrastructure with Claude Code as the unified scaffold (default system prompt overridden) for all models except GPT 5.4, which uses its native CodeX scaffold; results are averaged over 4 trials. \emph{Terminal-Bench 2.0} uses an 8 vCPU / 16\,GB sandbox with a 2\,hour wall-clock timeout, the Terminus-2 XML scaffold, and the verified 2.0 dataset (HuggingFace \path{zai-org/terminal-bench-2-verified}); 4 trials per model, baseline scores cited from official reports where not re-evaluated. \emph{MLE Bench Lite} runs each of the 22 competitions in a single-A30 sandbox for 24\,hours under our internal self-evolution scaffold (Bash + WebSearch); per-task scoring takes the best validation checkpoint and reports test-set medal rate; the final number is the mean of 3 independent 24-hour trials. \emph{VIBE-Pro} uses Claude Code as the verifier for both interaction logic and visual fidelity, end-to-end through a containerized deployment, 3 trials averaged. \emph{HyperTask}, \emph{MM Claw}, \emph{MEWC v2}, and \emph{Finance Modeling Pro} use expert-defined rubrics scored over 3 trials. \emph{BrowseComp}, \emph{Wide Search}, and \emph{RISE} share the \emph{WebExplorer}~\citep{liu2025webexplorer} agent framework, with light edits to the system prompt and tool descriptions; when token usage exceeds 30\,\% of the maximum context, all assistant replies and tool returns are dropped to keep the search alive. \emph{RISE} additionally enables a Playwright-based browser tool. \emph{GDPval-AA} is the Artificial Analysis re-evaluation on OpenAI's open GDPval dataset. \emph{AIME 2026}, \emph{GPQA-Diamond}, \emph{SciCode}, \emph{IFBench}, \emph{AA-LCR}, \emph{HLE}, and \emph{MMLU-Pro} are evaluated under the \emph{Artificial Analysis} Index v4.0 protocol with no tools and a single sample (pass@1).

\subsection{Main Results}

Table~\ref{tab:m2-main-results} reports M2.7 against four closed-weight frontier baselines (Claude Opus 4.6, Claude Sonnet 4.6, GPT 5.4, Gemini 3.1 Pro). The previous public M2 release (M2.5) is included as the within-series reference. The strongest score per row is \textbf{bolded}; ``--'' indicates that the model has not reported a score under our scaffold or has not yet released that benchmark at the time of writing.

\begin{table}[!t]
\scriptsize
\renewcommand{\arraystretch}{1.05}
\centering
\caption{\textbf{Performance of MiniMax-M2.7 versus closed-weight frontier baselines.} The MiniMax-M2.5 column gives the previous public release for within-series comparison.}
\setlength{\tabcolsep}{3.5pt}
\begin{tabular}{c|l|cc|cccc}
\toprule
& \multirow{2}{*}{\textbf{Benchmark}}
& \multicolumn{2}{c|}{\textbf{Ours}}
& \multicolumn{4}{c}{\textbf{Closed-weight frontier}} \\
\cmidrule(lr){3-4} \cmidrule(lr){5-8}
&
& \textbf{M2.7} & \textbf{M2.5}
& \makecell{\textbf{Opus} \\ \textbf{4.6}}
& \makecell{\textbf{Sonnet} \\ \textbf{4.6}}
& \makecell{\textbf{GPT} \\ \textbf{5.4}}
& \makecell{\textbf{Gemini} \\ \textbf{3.1 Pro}} \\
\midrule
\multirow{6}{*}{\rotatebox[origin=c]{90}{\shortstack{\textbf{Coding} \\ \textbf{Agent}}}}
 & SWE-bench Pro          & 56.2 & 55.4 & 57.3 & 57.2 & \textbf{57.7} & 54.2 \\
 & SWE-bench Multilingual & 76.5 & 74.1 & \textbf{77.8} & 75.9 & 70.5 & --   \\
 & Multi-SWE-bench        & \textbf{52.7} & 51.3 & 50.3 & 51.0 & 49.0 & --   \\
 & NL2Repo                & 39.8 & 26.6 & 43.7 & 43.3 & \textbf{46.8} & 35.9 \\
 & Terminal-Bench 2.0     & 57.0 & 51.7 & 65.4 & 59.1 & \textbf{75.1} & 68.5 \\
 & MLE Bench Lite         & 66.6 & 51.5 & \textbf{75.7} & 72.7 & 71.2 & 66.6 \\
\cmidrule(lr){2-8}
\multirow{2}{*}{\rotatebox[origin=c]{90}{\shortstack{\textbf{App} \\ \textbf{Dev}}}}
 & VIBE-Pro               & 55.6 & 54.2 & 55.6 & \textbf{56.1} & --   & 41.0 \\
 & HyperTask              & 67.6 & 59.4 & \textbf{75.7} & 74.1 & --   & 50.9 \\
\cmidrule(lr){2-8}
\multirow{3}{*}{\rotatebox[origin=c]{90}{\textbf{Search}}}
 & BrowseComp             & 77.8 & 76.3 & 84.0 & 74.7 & 82.7 & \textbf{85.9} \\
 & Wide Search            & 75.2 & 70.3 & \textbf{79.4} & 75.8 & 77.9 & --   \\
 & RISE                   & 64.3 & 50.2 & \textbf{68.5} & 58.8 & 63.3 & --   \\
\cmidrule(lr){2-8}
\multirow{5}{*}{\rotatebox[origin=c]{90}{\shortstack{\textbf{Office} \\ \textbf{\& Tools}}}}
 & GDPval-AA              & 50.0 & 35.0 & 55.0 & 57.0 & \textbf{58.0} & 41.0 \\
 & Toolathlon             & 46.3 & 38.3 & 47.2 & 44.8 & \textbf{54.6} & 48.8 \\
 & MM Claw                & 62.7 & 57.6 & \textbf{75.4} & 64.2 & 73.6 & 61.8 \\
 & MEWC v2                & 63.3 & 49.8 & 62.0 & \textbf{77.2} & 76.5 & 42.2 \\
 & Finance Modeling Pro   & 57.0 & 33.8 & 69.0 & 66.2 & \textbf{75.3} & 35.6 \\
\cmidrule(lr){2-8}
\multirow{7}{*}{\rotatebox[origin=c]{90}{\shortstack{\textbf{Reasoning} \\ \textbf{\& Knowledge}}}}
 & AIME 2026              & 94.2 & 87.2 & 92.5 & 92.7 & \textbf{97.0} & 88.7 \\
 & GPQA-Diamond           & 89.8 & 85.2 & 89.6 & 87.5 & 92.0 & \textbf{94.1} \\
 & SciCode                & 47.0 & 43.0 & 51.9 & 46.8 & 56.6 & \textbf{58.9} \\
 & IFBench                & 76.0 & 72.0 & 53.1 & 56.6 & 73.9 & \textbf{77.1} \\
 & AA-LCR                 & 72.0 & 65.0 & 70.7 & 70.7 & \textbf{74.0} & 72.7 \\
 & HLE                    & 28.0 & 19.0 & 36.7 & 30.0 & 41.6 & \textbf{44.7} \\
 & MMLU-Pro               & 81.8 & 85.2 & 89.1 & 87.3 & 87.5 & \textbf{91.2} \\
\bottomrule
\end{tabular}
\label{tab:m2-main-results}
\end{table}

\noindent \textbf{Software engineering and coding agent.} M2.7 is broadly competitive across the agentic-coding suite. It scores 56.2 on \emph{SWE-bench Pro}, 76.5 on \emph{SWE-bench Multilingual}, takes the top score among compared models on \emph{Multi-SWE-bench} at 52.7, and reaches 57.0 on \emph{Terminal-Bench 2.0}. On \emph{NL2Repo} M2.7 reaches 39.8, a 13-point jump from M2.5 (26.6) that reflects the new full-stack repository data introduced in \S\ref{sec:agentic-coding-data}. On \emph{MLE Bench Lite} M2.7 reaches a 66.6\,\% medal rate, a 15-point absolute jump over M2.5; we discuss this case in detail in \S\ref{sec:eval-case}.

\noindent \textbf{Application development.} On \emph{VIBE-Pro} M2.7 reaches 55.6, on par with the leading closed-weight baselines. On the harder long-horizon \emph{HyperTask} suite, M2.7 reaches 67.6, an 8-point improvement over M2.5. Application development is one of the cleanest domains for the M2 design thesis: with $\sim$10\,B activated parameters, the AppDev-targeted training data and Agent-as-a-Verifier reward (\S\ref{sec:agentic-coding-data}) close most of the gap to frontier models that activate an order of magnitude more parameters per token.

\noindent \textbf{Cowork --- search and deep research.} On the open-web search-and-synthesize triad, M2.7 reaches 77.8 on \emph{BrowseComp}, 75.2 on \emph{Wide Search}, and 64.3 on our internal \emph{RISE} benchmark (designed to require non-trivial multi-step browsing and cross-source corroboration with a Playwright browser tool). RISE also marks the largest within-series gain in this block, climbing 14 points from M2.5 (50.2). M2.7 is most competitive on tasks demanding longer planning horizons and richer web interaction, consistent with the verifier-grounded data pipeline used to construct our deep-search training corpus (\S\ref{sec:agentic-cowork-data}) and the persistent reasoning state of interleaved thinking across many turns (\S\ref{sec:interleaved-thinking}).

\noindent \textbf{Cowork --- agent, office, and workspace.} On heterogeneous tool-use benchmarks, M2.7 reaches 50.0 on \emph{GDPval-AA} and 46.3 on \emph{Toolathlon}. On Excel-grounded operation, M2.7 reaches 63.3 on \emph{MEWC v2} and 57.0 on \emph{Finance Modeling Pro}; on \emph{MM Claw} M2.7 reaches 62.7. This block shows the largest within-series headroom of any capability area in the M2 series, with substantial M2.5\,$\to$\,M2.7 gains on Excel-grounded benchmarks (\emph{MEWC v2} $+13.5$, \emph{Finance Modeling Pro} $+23.2$) and on \emph{GDPval-AA} ($+15.0$).

\noindent \textbf{Reasoning and knowledge.} M2.7 is competitive on the reasoning-heavy benchmarks where evaluation reduces to a single sample without tools. M2.7 scores 94.2 on \emph{AIME 2026}, 89.8 on \emph{GPQA-Diamond}, 76.0 on \emph{IFBench}, and 72.0 on \emph{AA-LCR}, placing it in the frontier band on each. The \emph{IFBench} result in particular reflects the multi-domain RL strategy and rubric-based response filtering described in \S\ref{sec:reasoning-data} and \S\ref{sec:rl}. On the broader knowledge benchmarks, M2.7 reaches 81.8 on \emph{MMLU-Pro}, 47.0 on \emph{SciCode}, and 28.0 on \emph{HLE}.

\begin{figure*}[!t]
\centering
\includegraphics[width=\textwidth]{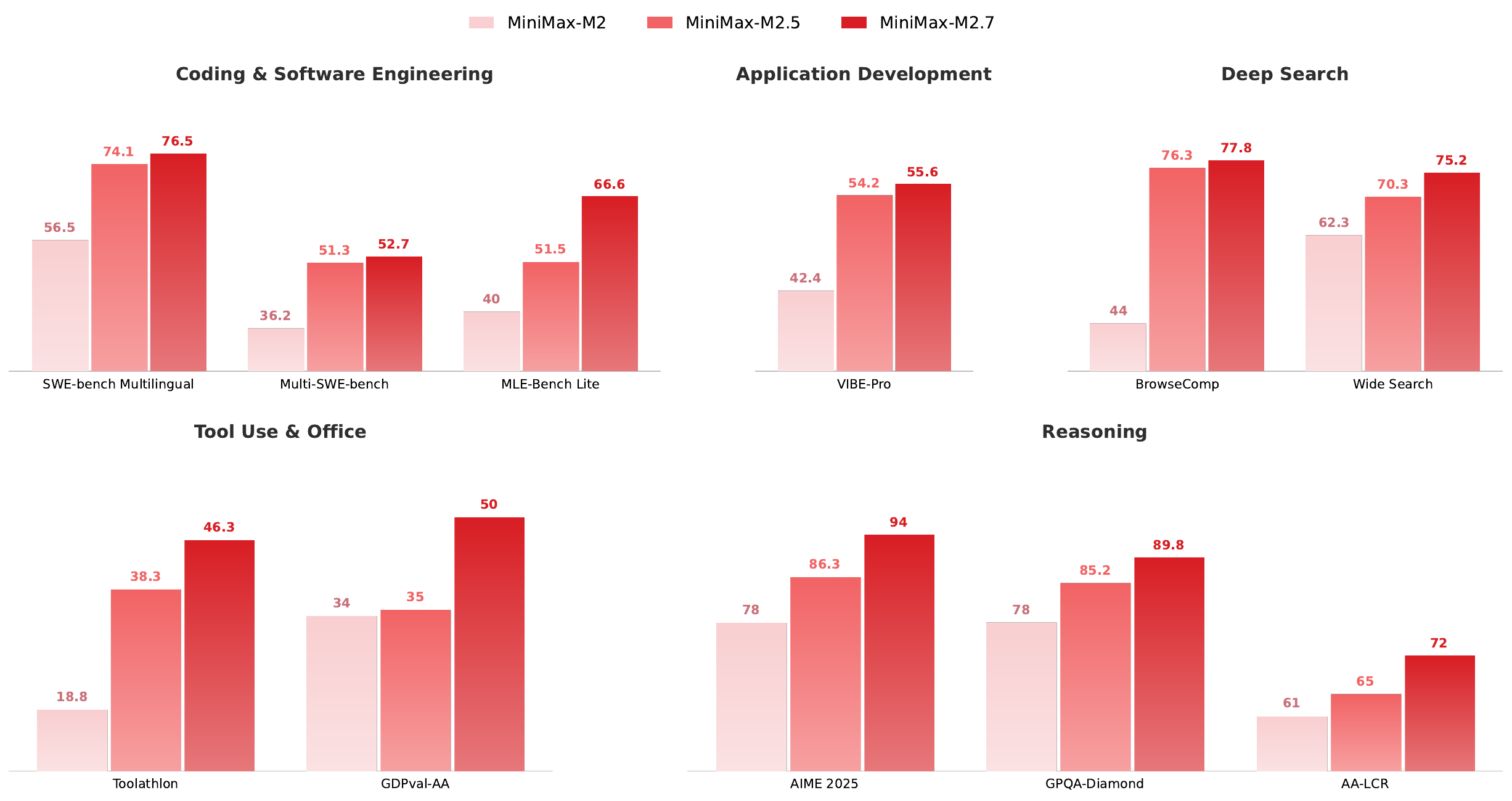}
\caption{\textbf{Capability progression of the MiniMax-M2 series across eleven benchmarks.} Each panel groups benchmarks by capability area, and within each benchmark the three bars correspond to the original M2 release, M2.5, and the current M2.7. AIME numbers in this figure refer to AIME 2025, the only AIME edition reported for all three checkpoints; the BrowseComp number for M2 is the Hugging Face official score. All eleven benchmarks improve across the three checkpoints, with the largest gains concentrated in the agentic deep-search, tool-use, and autonomous ML-engineering domains where the M2.5 / M2.7 data pipelines added new task families.}
\label{fig:m2-progression}
\end{figure*}

\noindent \textbf{Within-series progression.} To complement the cross-model snapshot of Table~\ref{tab:m2-main-results}, Figure~\ref{fig:m2-progression} traces the trajectory of the M2 series itself from the original M2 release through M2.5 to the current M2.7, restricted to the eleven benchmarks for which all three checkpoints have been evaluated under our scaffold. Two patterns stand out. First, every benchmark in this set improves across the three checkpoints, with absolute gains ranging from $+11$ points (AA-LCR, GPQA-Diamond) to $+33.8$ points (BrowseComp). Second, the size of the gain tracks the data-pipeline investments described in \S\ref{sec:post-training-data}: the benchmarks where the M2.5 / M2.7 corpora introduced new task families---deep search (BrowseComp $+33.8$, Wide Search $+12.9$), tool use (Toolathlon $+27.5$, GDPval-AA $+16.0$), and autonomous ML engineering (MLE Bench Lite $+26.6$)---show the steepest jumps, while benchmarks the original M2 was already strong on (SWE-bench Multilingual, Multi-SWE-bench) progress more incrementally. Reasoning benchmarks (AIME 2025 $+16.0$, GPQA-Diamond $+11.8$, AA-LCR $+11.0$) follow a steadier curve consistent with the multi-axis scaling of the reasoning data pipeline. Taken together, the figure indicates that each of the three contribution axes of the M2 series---agentic data, the Forge RL system (\S\ref{sec:rl}), and self-evolution (\S\ref{sec:self-evolution})---translates into a stable improvement trajectory at every release rather than concentrating in any single checkpoint.

\subsection{Case Study: Self-Evolution on MLE Bench Lite}
\label{sec:eval-case}

\noindent The MLE Bench Lite result in Table~\ref{tab:m2-main-results} is the most direct evidence for the self-evolution capability whose system design is described in \Cref{sec:self-evolution}. The underlying driver is M2.7's strength in Machine Learning Engineering (MLE): on OpenAI's \emph{MLE Bench Lite}~\citep{chan2025mlebench}, M2.7 ties Gemini 3.1 Pro, demonstrating the frontier-level ability required to independently orchestrate ML pipelines and modify its own training scaffolds.

\noindent To test this capability under a controlled setting, we evaluated M2.7 as an independent ML engineer across 22 competitions from \emph{MLE Bench Lite}. While computationally lightweight, these tasks cover all stages of a standard ML workflow.

\noindent To guide the model, we implemented a simple autonomous harness driven by short-term memory and self-feedback, with no human-written code in the harness itself. After completing an iteration, the agent documents a memory file and performs rigorous self-criticism. This self-reflective critique establishes explicit optimization directions for subsequent runs, allowing the model to build upon an accumulated feedback chain.

\begin{figure}[h]
\centering
\includegraphics[width=0.74\textwidth]{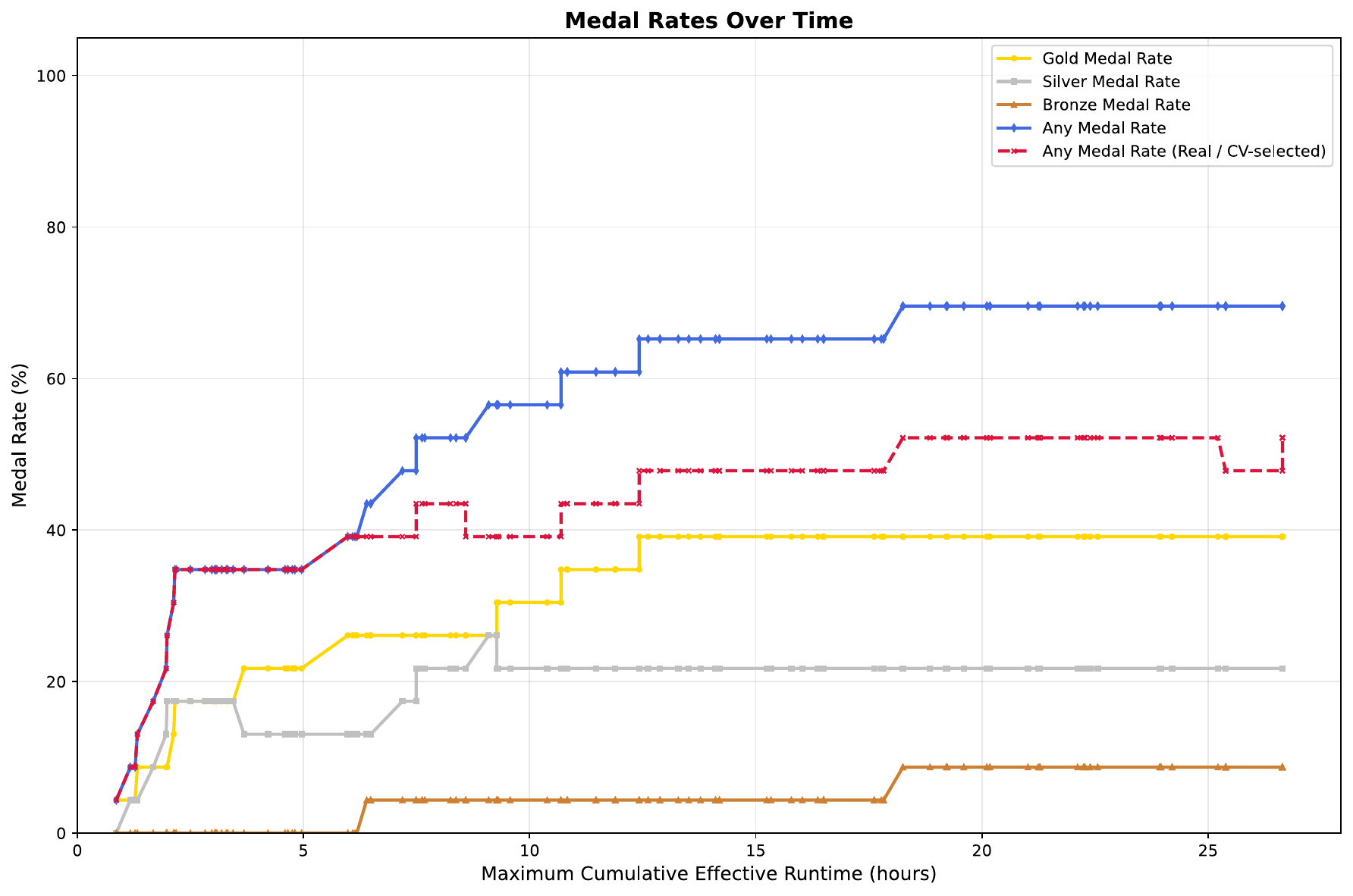}
\caption{Medal rate of M2.7 on \emph{MLE Bench Lite} across iterative trials.}
\label{fig:mle-medal-rate}
\end{figure}

\noindent We executed three independent trials, allowing 24\,hours of iterative evolution per run. As illustrated in Figure~\ref{fig:mle-medal-rate}, M2.7 demonstrated clear cumulative improvement, steadily increasing its medal rate over time. The best run yielded 9 gold medals, 5 silver medals, and 1 bronze medal. Averaging a 66.6\% medal rate across trials, M2.7 ties Gemini 3.1 Pro, demonstrating its capacity to autonomously navigate and optimize complex, end-to-end ML pipelines.

\noindent We highlight this case not just for the numerical outcome but for the qualitative observation that M2.7 willingly debugs its own training scaffold, modifies configuration files, and iterates over hundreds of rounds---behaviors that close the \emph{mini-activations $\to$ max real-world intelligence} loop the M2 series is designed around.

%% file: conclusion.tex
\section{Conclusion}

We presented the \textbf{MiniMax-M2 series}, a family of Mixture-of-Experts language models built around the thesis that \emph{mini activations can unleash maximum real-world intelligence}. The flagship M2 pairs a 9.8B-activated / 229.9B-total backbone with three components that co-evolve from M2 through M2.5 to the current M2.7 checkpoint: agent-driven data pipelines that ground every training trajectory in an executable workspace and an artifact-aligned reward; Forge, an agent-native RL system that scales long-horizon training across both white-box and black-box agent loops; and an early operational form of self-evolution, in which M2.7 autonomously debugs its own training runs and modifies its agent scaffold. Together, these components translate a $\sim$10\,B activated-parameter footprint into parity with frontier systems an order of magnitude larger in per-step compute on agentic coding, agentic cowork, and reasoning \& knowledge benchmarks. We view these results as one step along a longer trajectory: each axis---data, RL system, and self-evolution---remains far from saturation, and subsequent M2.x checkpoints will continue scaling all three in concert.

%% file: app.tex
\appendix
\input{showcase_format}
\newpage

\section{Contributors}
The contributors to the report are listed in alphabetical order as follows:

Aili Chen,
Aonian Li,
Baichuan Zhou,
Bangwei Gong,
Binyang Jiang,
Boji Dan,
Changhao Zhang,
Changqing Yu,
Chao Wang,
Cheng Ma,
Cheng Zhong,
Cheng Zhu,
Chengjun Xiao,
Chengyi Yang,
Chengyu Du,
Chenyang Zhang,
Chi Zhang,
Chuangyi Huang,
Chunhao Zhang,
Chunhui Du,
Chunyu Zhao,
Congchao Guo,
Da Chen,
Deming Ding,
Dianjun Sun,
Dong Li,
Dongyu Zhang,
Enhui Yang,
Fei Yu,
Guang Zheng,
Guodong Zheng,
Guohong Li,
Haichao Zhu,
Haigang Zhou,
Haimo Zhang,
Han Ding,
Hao Zhang,
Haohai Sun,
Haolin Lyu,
Haonan Lu,
Haoyu Wang,
Huajie Shi,
Huiyang Li,
Jiacheng Chen,
Jian Zhang,
Jiaqi Zhuang,
Jiaren Cai,
Jiaxin Pan,
Jiayao Li,
Jiayuan Song,
Jichuan Zhang,
Jie Wang,
Jihao Gu,
Jin Zhu,
Jingwei Dong,
Jingyang Li,
Jingyu Zhang,
Jingze Zhuang,
Jinhao Tian,
Jinli Liu,
Jinyi Hu,
Jun Tao,
Jun Zhang,
Junbin Ruan,
Junhao Xu,
Junjie Yan,
Junteng Liu,
Junxian He,
Kang Xu,
Ke Ji,
Ke Yang,
Kecheng Xiao,
Keyu Duan,
Keyu Li,
Le Han,
Letian Ruan,
Li Yuan,
Lianfei Yu,
Liheng Feng,
Lijie Mo,
Lin Li,
Linge Du,
Lingye Bao,
Lingyu Yang,
Lingyuan Zhou,
Loki,
Lu Chen,
Lunbin Zeng,
Ming Li,
Ming Zhong,
Mingliang Tao,
Mingyuan Chi,
Mujie Lin,
Nan Hu,
Ningxin Chen,
Peiyin Zhu,
Peng Gao,
Pengcheng Gao,
Pengfei Li,
Penglin Li,
Pengyu Zhao,
Qibin Ren,
Qibing Ren,
Qidi Xu,
Qihan Ren,
Qile Li,
Qin Wang,
Quanliang Chen,
Qunhong Zeng,
Rong Tian,
Rongxin Guo,
Rui Dong,
Ruitao Leng,
Ruize Zhang,
Shanqi Liu,
Shaoxiang Chen,
Shaoyu Chen,
Sheng Jia,
Shun Yao,
Shuoran Zhao,
Shuqi Yu,
Sichen Li,
Sicheng Pan,
Songquan Zhu,
Tengfei Li,
Tian Xie,
Tiancheng Qin,
Tianle Li,
Tianrun Liang,
Wei Liu,
Weiqi Xu,
Weitao Li,
Weixiang Chen,
Weiyu Cheng,
Weiyu Zhang,
Wenhu Chen,
Wenqian Zhao,
Xiancai Chen,
Xiangjun Song,
Xiangyuan Wang,
Xianzhen Luo,
Xiao Luo,
Xiao Su,
Xiaobo Li,
Xiaodong Han,
Xiaojie Wu,
Xihao Song,
Xingyi Han,
Xinyu Guan,
Xuan Lu,
Xun Zou,
Xunhao Lai,
Xutong Li,
Xuyang Shen,
Yan Gong,
Yan Ma,
Yang Jiao,
Yang Wang,
Yang Xu,
Yangsen Wang,
Ye Tang,
Yicheng Chen,
Yihang Wang,
Yinran Qiu,
Yiqi Shi,
Yiting Guo,
Yiwen Huang,
Yixuan Wang,
Yongyi Hu,
Yu Gao,
Yu Zhang,
Yuan Li,
Yuanxiang Ying,
Yuanzhen Zhang,
Yubo Wang,
Yuchen Song,
Yufeng Yang,
Yuhang Meng,
Yuhang Miao,
Yuhao Li,
Yujie Liu,
Yulin Hu,
Yunan Huang,
Yunji Li,
Yunyi Huang,
Yusen Zhang,
Yusu Hong,
Yutao Xie,
Yutong Zhang,
Yuwen Liao,
Yuxuan Shi,
Yuze Wenren,
Zebin Li,
Zehan Li,
Zejian Luo,
Zeyu Jin,
Zeyuan Sun,
Zhanpeng Zhou,
Zhaochen Su,
Zhendong Li,
Zhengmao Zhu,
Zhengyuan Peng,
Zhenhua Fan,
Zhi Zhang,
Zhichao Xu,
Zhiheng Lv,
Zhikang Xu,
Zhitao He,
Zhiwei He,
Zhongyuan Li,
Zibo Gao,
Zijia Wu,
Zijian Song,
Zijian Zhou,
Zijun Sun,
Zishan Huang,
Ziying Chen,
Ziyue Ge

%% file: M2_cite.bib
@inproceedings{dai2024deepseekmoe,
  title     = {DeepSeekMoE: Towards Ultimate Expert Specialization in Mixture-of-Experts Language Models},
  author    = {Dai, Damai and Deng, Chengqi and Zhao, Chenggang and Xu, R. X. and Gao, Huazuo and Chen, Deli and Li, Jiashi and Zeng, Wangding and Yu, Xingkai and Wu, Y. and Xie, Zhenda and Li, Y. K. and Huang, Panpan and Zhao, Fuli and Sun, Xiaodong and Liu, Aixin and Liang, Wenfeng},
  booktitle = {Proceedings of the 62nd Annual Meeting of the Association for Computational Linguistics (Volume 1: Long Papers)},
  year      = {2024}
}

@misc{wang2024auxfree,
  title         = {Auxiliary-Loss-Free Load Balancing Strategy for Mixture-of-Experts},
  author        = {Wang, Lean and Gao, Huazuo and Zhao, Chenggang and Sun, Xu and Dai, Damai},
  year          = {2024},
  eprint        = {2408.15664},
  archivePrefix = {arXiv},
  primaryClass  = {cs.CL}
}

@inproceedings{ainslie2023gqa,
  title     = {{GQA}: Training Generalized Multi-Query Transformer Models from Multi-Head Checkpoints},
  author    = {Ainslie, Joshua and Lee-Thorp, James and de Jong, Michiel and Zemlyanskiy, Yury and Lebr{\'o}n, Federico and Sanghai, Sumit},
  booktitle = {Proceedings of the 2023 Conference on Empirical Methods in Natural Language Processing},
  year      = {2023}
}

@article{su2024roformer,
  title   = {{RoFormer}: Enhanced transformer with Rotary Position Embedding},
  author  = {Su, Jianlin and Ahmed, Murtadha and Lu, Yu and Pan, Shengfeng and Bo, Wen and Liu, Yunfeng},
  journal = {Neurocomputing},
  year    = {2024},
  volume  = {568}
}

@inproceedings{gloeckle2024better,
  title     = {Better \& Faster Large Language Models via Multi-Token Prediction},
  author    = {Gloeckle, Fabian and Idrissi, Badr Youbi and Rozi{\`e}re, Baptiste and Lopez-Paz, David and Synnaeve, Gabriel},
  booktitle = {Proceedings of the 41st International Conference on Machine Learning},
  year      = {2024}
}

@misc{deepseekai2024v3,
  title         = {{DeepSeek-V3} Technical Report},
  author        = {{DeepSeek-AI}},
  year          = {2024},
  eprint        = {2412.19437},
  archivePrefix = {arXiv},
  primaryClass  = {cs.CL}
}

@inproceedings{leviathan2023fast,
  title     = {Fast Inference from Transformers via Speculative Decoding},
  author    = {Leviathan, Yaniv and Kalman, Matan and Matias, Yossi},
  booktitle = {Proceedings of the 40th International Conference on Machine Learning},
  year      = {2023}
}

@misc{beltagy2020longformer,
  title         = {Longformer: The Long-Document Transformer},
  author        = {Beltagy, Iz and Peters, Matthew E. and Cohan, Arman},
  year          = {2020},
  eprint        = {2004.05150},
  archivePrefix = {arXiv},
  primaryClass  = {cs.CL}
}

@inproceedings{qin2024lightning,
  title     = {Lightning Attention-2: A Free Lunch for Handling Unlimited Sequence Lengths in Large Language Models},
  author    = {Qin, Zhen and Sun, Weigao and Li, Dong and Shen, Xuyang and Sun, Weixuan and Zhong, Yiran},
  booktitle = {Proceedings of the 41st International Conference on Machine Learning},
  year      = {2024}
}

@misc{minimax2025minimax01,
  title         = {{MiniMax-01}: Scaling Foundation Models with Lightning Attention},
  author        = {{MiniMax}},
  year          = {2025},
  eprint        = {2501.08313},
  archivePrefix = {arXiv},
  primaryClass  = {cs.CL}
}

@misc{olsson2022induction,
  title         = {In-context Learning and Induction Heads},
  author        = {Olsson, Catherine and Elhage, Nelson and Nanda, Neel and Joseph, Nicholas and DasSarma, Nova and Henighan, Tom and Mann, Ben and Askell, Amanda and Bai, Yuntao and Chen, Anna and others},
  year          = {2022},
  eprint        = {2209.11895},
  archivePrefix = {arXiv},
  primaryClass  = {cs.LG}
}

@misc{wu2024retrieval,
  title         = {Retrieval Head Mechanistically Explains Long-Context Factuality},
  author        = {Wu, Wenhao and Wang, Yizhong and Xiao, Guangxuan and Peng, Hao and Fu, Yao},
  year          = {2024},
  eprint        = {2404.15574},
  archivePrefix = {arXiv},
  primaryClass  = {cs.CL}
}

@misc{minimax2025m1,
  title         = {{MiniMax-M1}: Scaling Test-Time Compute Efficiently with Lightning Attention},
  author        = {{MiniMax}},
  year          = {2025},
  eprint        = {2506.13585},
  archivePrefix = {arXiv},
  primaryClass  = {cs.CL}
}

@article{chen2026dive,
  title={DIVE: Scaling Diversity in Agentic Task Synthesis for Generalizable Tool Use},
  author={Chen, Aili and Zhang, Chi and Liu, Junteng and Chen, Jiangjie and Du, Chengyu and Li, Yunji and Zhong, Ming and Wang, Qin and Zhu, Zhengmao and Song, Jiayuan and others},
  journal={arXiv preprint arXiv:2603.11076},
  year={2026}
}

@article{liu2025webexplorer,
  title={Webexplorer: Explore and evolve for training long-horizon web agents},
  author={Liu, Junteng and Li, Yunji and Zhang, Chi and Li, Jingyang and Chen, Aili and Ji, Ke and Cheng, Weiyu and Wu, Zijia and Du, Chengyu and Xu, Qidi and others},
  journal={arXiv preprint arXiv:2509.06501},
  year={2025}
}

@misc{luo2026cvefactory,
  title         = {{CVE-Factory}: Scaling Expert-Level Agentic Tasks for Code Security Vulnerability},
  author        = {Luo, Xianzhen and Zhang, Jingyuan and Zhou, Shiqi and Huang, Rain and Xiao, Chuan and Zhu, Qingfu and Ma, Zhiyuan and Yue, Xing and Yue, Yang and Zeng, Wencong and Che, Wanxiang},
  year          = {2026},
  eprint        = {2602.03012},
  archivePrefix = {arXiv},
  primaryClass  = {cs.CR}
}

@misc{minimax2026deepdive,
  author       = {{MiniMax AI}},
  title        = {A Deep Dive into the MiniMax-M2-her},
  howpublished = {\url{https://x.com/MiniMax_AI/status/2016521115355799660}},
  year         = {2026}
}

@misc{yang2024swesmith,
  title         = {{SWE-smith}: Scaling Data for Software Engineering Agents},
  author        = {Yang, John and Lee, Kilian and Lieret, Kilian and Yao, Shunyu and Xie, Yujia and Wettig, Alexander and Liu, Ofir Press Yao and Jimenez, Carlos E. and Khattab, Omar and Yan, Sean and others},
  year          = {2024},
  eprint        = {2504.21798},
  archivePrefix = {arXiv},
  primaryClass  = {cs.SE}
}

@misc{openai2025gpt5,
  title        = {Introducing {GPT-5}},
  author       = {{OpenAI}},
  year         = {2025},
  howpublished = {\url{https://openai.com/gpt-5/}}
}

@misc{anthropic2025claude46,
  title        = {{Claude} {Opus} 4.6 and {Sonnet} 4.6 Model Card},
  author       = {{Anthropic}},
  year         = {2025},
  howpublished = {\url{https://www.anthropic.com/news/claude-4-6}}
}

@misc{google2025gemini31,
  title        = {{Gemini} 3.1 Pro},
  author       = {{Google DeepMind}},
  year         = {2025},
  howpublished = {\url{https://deepmind.google/technologies/gemini/}}
}

@inproceedings{jimenez2024swebench,
  title     = {{SWE}-bench: Can Language Models Resolve Real-World {GitHub} Issues?},
  author    = {Jimenez, Carlos E. and Yang, John and Wettig, Alexander and Yao, Shunyu and Pei, Kexin and Press, Ofir and Narasimhan, Karthik},
  booktitle = {International Conference on Learning Representations},
  year      = {2024}
}

@misc{swebenchpro2025,
  title         = {{SWE-bench Pro}: Industry-Grade Repository-Level Software Engineering Benchmark},
  author        = {Wang, Yuxuan and others},
  year          = {2025},
  eprint        = {2509.16941},
  archivePrefix = {arXiv},
  primaryClass  = {cs.SE}
}

@misc{rashid2025swebenchmultilingual,
  title         = {{SWE-bench Multilingual}: Evaluating Coding Agents Beyond Python},
  author        = {Rashid, Layla El and others},
  year          = {2025},
  eprint        = {2504.02605},
  archivePrefix = {arXiv},
  primaryClass  = {cs.SE}
}

@misc{multiswebench2025,
  title         = {{Multi-SWE-bench}: A Multilingual Benchmark for Issue Resolving},
  author        = {Zan, Daoguang and others},
  year          = {2025},
  eprint        = {2504.02605},
  archivePrefix = {arXiv},
  primaryClass  = {cs.SE}
}

@inproceedings{merrill2026terminalbench,
  title     = {Terminal-Bench: Benchmarking Agents on Hard, Realistic Tasks in Command Line Interfaces},
  author    = {Merrill, Mike A. and Shaw, Alexander Glenn and Carlini, Nicholas and others},
  booktitle = {International Conference on Learning Representations},
  year      = {2026}
}

@misc{chan2025mlebench,
  title         = {{MLE-bench}: Evaluating Machine Learning Agents on Machine Learning Engineering},
  author        = {Chan, Jun Shern and Chowdhury, Neil and Jaffe, Oliver and Aung, James and Sherburn, Dane and Mays, Evan and Starace, Giulio and Liu, Kevin and Maksin, Leon and Patwardhan, Tejal and Weng, Lilian and Madry, Aleksander},
  year          = {2025},
  eprint        = {2410.07095},
  archivePrefix = {arXiv},
  primaryClass  = {cs.CL}
}

@misc{wei2025browsecomp,
  title         = {{BrowseComp}: A Simple Yet Challenging Benchmark for Browsing Agents},
  author        = {Wei, Jason and Sun, Yifan and Karina, Anastasia and Patwardhan, Tejal and Chen, Mark and Glaese, Amelia},
  year          = {2025},
  eprint        = {2504.12516},
  archivePrefix = {arXiv},
  primaryClass  = {cs.CL}
}

@misc{liu2025widesearch,
  title         = {{Wide Search}: Benchmarking Long-Horizon Multi-Source Web Research},
  author        = {Liu, Hongru and others},
  year          = {2025},
  eprint        = {2510.05746},
  archivePrefix = {arXiv},
  primaryClass  = {cs.CL}
}

@misc{huang2025toolathlon,
  title         = {{Toolathlon}: A Heterogeneous Tool-Use Benchmark for {LLM} Agents},
  author        = {Huang, Yufei and others},
  year          = {2025},
  eprint        = {2510.04534},
  archivePrefix = {arXiv},
  primaryClass  = {cs.CL}
}

@misc{patwardhan2025gdpvalevaluatingaimodel,
  title         = {{GDPval}: Evaluating {AI} Model Performance on Real-World Economically Valuable Tasks},
  author        = {Patwardhan, Tejal and Dias, Rachel and Proehl, Elizabeth and Kim, Grace and Wang, Michele and Watkins, Olivia and Fishman, Sim{\'o}n Posada and Aljubeh, Marwan and Thacker, Phoebe and Fauconnet, Laurance and Kim, Natalie S. and Chao, Patrick and Miserendino, Samuel and Chabot, Gildas and Li, David and Sharman, Michael and Barr, Alexandra and Glaese, Amelia and Tworek, Jerry},
  year          = {2025},
  eprint        = {2510.04374},
  archivePrefix = {arXiv},
  primaryClass  = {cs.LG}
}

@misc{aime2026,
  title        = {American Invitational Mathematics Examination 2026},
  author       = {{Mathematical Association of America}},
  year         = {2026},
  howpublished = {\url{https://www.maa.org/math-competitions/aime}}
}

@inproceedings{rein2024gpqa,
  title     = {{GPQA}: A Graduate-Level Google-Proof {Q\&A} Benchmark},
  author    = {Rein, David and Hou, Betty Li and Stickland, Asa Cooper and Petty, Jackson and Pang, Richard Yuanzhe and Dirani, Julien and Michael, Julian and Bowman, Samuel R.},
  booktitle = {Conference on Language Modeling},
  year      = {2024}
}

@misc{tian2024scicode,
  title         = {{SciCode}: A Research Coding Benchmark Curated by Scientists},
  author        = {Tian, Minyang and Gao, Luyu and Zhang, Shizhuo Dylan and Chen, Xinan and Fan, Cunwei and Guo, Xuefei and Haas, Roland and Ji, Pan and Krongchon, Kittithat and Li, Yao and Liu, Shengyan and Luo, Di and Ma, Yutao and Tong, Hao and Trinh, Kha and Tian, Chenyu and Tian, Zihan and Wang, Yufeng and Wu, Xinyuan and You, Hao Tianci and Yuan, Boxuan and Zhang, Qihang and Zhao, Lvyou and Zhao, Yingdong and Zhao, Yujie and Zhao, Zhiwei and Wang, Yu and Liu, Tina and Liu, Hongru and Eckart, Curtis and others},
  year          = {2024},
  eprint        = {2407.13168},
  archivePrefix = {arXiv},
  primaryClass  = {cs.AI}
}

@misc{pyatkin2025ifbench,
  title         = {{IFBench}: Probing Verifiable Instruction Following on Hard Constraints},
  author        = {Pyatkin, Valentina and others},
  year          = {2025},
  eprint        = {2507.02833},
  archivePrefix = {arXiv},
  primaryClass  = {cs.CL}
}

@misc{phan2025humanity,
  title         = {Humanity's Last Exam},
  author        = {Phan, Long and Gatti, Alice and Han, Ziwen and Li, Nathaniel and Hu, Josephina and others},
  year          = {2025},
  eprint        = {2501.14249},
  archivePrefix = {arXiv},
  primaryClass  = {cs.CL}
}

@inproceedings{wang2024mmlu,
  title     = {{MMLU-Pro}: A More Robust and Challenging Multi-Task Language Understanding Benchmark},
  author    = {Wang, Yubo and Ma, Xueguang and Zhang, Ge and Ni, Yuansheng and Chandra, Abhranil and Guo, Shiguang and Ren, Weiming and Arulraj, Aaran and He, Xuan and Jiang, Ziyan and Li, Tianle and Ku, Max and Wang, Kai and Zhuang, Alex and Fan, Rongqi and Yue, Xiang and Chen, Wenhu},
  booktitle = {Advances in Neural Information Processing Systems},
  year      = {2024}
}

@inproceedings{hendrycks2021mmlu,
  title     = {Measuring Massive Multitask Language Understanding},
  author    = {Hendrycks, Dan and Burns, Collin and Basart, Steven and Zou, Andy and Mazeika, Mantas and Song, Dawn and Steinhardt, Jacob},
  booktitle = {International Conference on Learning Representations},
  year      = {2021}
}

@inproceedings{hendrycks2021math,
  title     = {Measuring Mathematical Problem Solving with the {MATH} Dataset},
  author    = {Hendrycks, Dan and Burns, Collin and Kadavath, Saurav and Arora, Akul and Basart, Steven and Tang, Eric and Song, Dawn and Steinhardt, Jacob},
  booktitle = {Proceedings of the Neural Information Processing Systems Track on Datasets and Benchmarks},
  year      = {2021}
}

@misc{chen2021humaneval,
  title         = {Evaluating Large Language Models Trained on Code},
  author        = {Chen, Mark and Tworek, Jerry and Jun, Heewoo and Yuan, Qiming and Pinto, Henrique Ponde de Oliveira and Kaplan, Jared and Edwards, Harri and Burda, Yuri and Joseph, Nicholas and Brockman, Greg and others},
  year          = {2021},
  eprint        = {2107.03374},
  archivePrefix = {arXiv},
  primaryClass  = {cs.LG}
}

@misc{clark2018arc,
  title         = {Think You Have Solved Question Answering? Try {ARC}, the {AI2} Reasoning Challenge},
  author        = {Clark, Peter and Cowhey, Isaac and Etzioni, Oren and Khot, Tushar and Sabharwal, Ashish and Schoenick, Carissa and Tafjord, Oyvind},
  year          = {2018},
  eprint        = {1803.05457},
  archivePrefix = {arXiv},
  primaryClass  = {cs.AI}
}

@inproceedings{suzgun2023challenging,
  title     = {Challenging {BIG-Bench} Tasks and Whether Chain-of-Thought Can Solve Them},
  author    = {Suzgun, Mirac and Scales, Nathan and Sch{\"a}rli, Nathanael and Gehrmann, Sebastian and Tay, Yi and Chung, Hyung Won and Chowdhery, Aakanksha and Le, Quoc V. and Chi, Ed H. and Zhou, Denny and Wei, Jason},
  booktitle = {Findings of the Association for Computational Linguistics: ACL 2023},
  year      = {2023}
}

@inproceedings{bai2024longbench,
  title     = {{LongBench}: A Bilingual, Multitask Benchmark for Long Context Understanding},
  author    = {Bai, Yushi and Lv, Xin and Zhang, Jiajie and Lyu, Hongchang and Tang, Jiankai and Huang, Zhidian and Du, Zhengxiao and Liu, Xiao and Zeng, Aohan and Hou, Lei and Dong, Yuxiao and Tang, Jie and Li, Juanzi},
  booktitle = {Proceedings of the 62nd Annual Meeting of the Association for Computational Linguistics (Volume 1: Long Papers)},
  year      = {2024}
}

@misc{yen2024helmet,
  title         = {{HELMET}: How to Evaluate Long-Context Language Models Effectively and Thoroughly},
  author        = {Yen, Howard and Gao, Tianyu and Hou, Minmin and Ding, Ke and Fleischer, Daniel and Izsak, Peter and Wasserblat, Moshe and Chen, Danqi},
  year          = {2024},
  eprint        = {2410.02694},
  archivePrefix = {arXiv},
  primaryClass  = {cs.CL}
}

@misc{hsieh2024ruler,
  title         = {{RULER}: What's the Real Context Size of Your Long-Context Language Models?},
  author        = {Hsieh, Cheng-Ping and Sun, Simeng and Kriman, Samuel and Acharya, Shantanu and Rekesh, Dima and Jia, Fei and Ginsburg, Boris},
  year          = {2024},
  eprint        = {2404.06654},
  archivePrefix = {arXiv},
  primaryClass  = {cs.CL}
}

@misc{tanzer2024mtob,
  title         = {A Benchmark for Learning to Translate a New Language from One Grammar Book},
  author        = {Tanzer, Garrett and Suzgun, Mirac and Visser, Eline and Jurafsky, Dan and Melas-Kyriazi, Luke},
  year          = {2024},
  eprint        = {2309.16575},
  archivePrefix = {arXiv},
  primaryClass  = {cs.CL}
}

@misc{chollet2019arcagi,
  title         = {On the Measure of Intelligence},
  author        = {Chollet, Fran{\c{c}}ois},
  year          = {2019},
  eprint        = {1911.01547},
  archivePrefix = {arXiv},
  primaryClass  = {cs.AI}
}

@inproceedings{mialon2024gaia,
  title     = {{GAIA}: A Benchmark for General {AI} Assistants},
  author    = {Mialon, Gr{\'e}goire and Fourrier, Cl{\'e}mentine and Wolf, Thomas and LeCun, Yann and Scialom, Thomas},
  booktitle = {International Conference on Learning Representations},
  year      = {2024}
}

@misc{chen2025xbench,
  title         = {xbench: Tracking Agents Productivity Scaling with Profession-Aligned Real-World Evaluations},
  author        = {Chen, Kaiyuan and others},
  year          = {2025},
  eprint        = {2506.13651},
  archivePrefix = {arXiv},
  primaryClass  = {cs.AI}
}

@misc{aime2025,
  title        = {American Invitational Mathematics Examination 2025},
  author       = {{Mathematical Association of America}},
  year         = {2025},
  howpublished = {\url{https://www.maa.org/math-competitions/aime}}
}

@misc{zhou2025browsecompzhbenchmarkingwebbrowsing,
  title         = {{BrowseComp-ZH}: Benchmarking Web Browsing Ability of Large Language Models in Chinese},
  author        = {Zhou, Peilin and Leon, Bruce and Ying, Xiang and Zhang, Can and Shao, Yifan and Ye, Qichen and Chong, Dading and Jin, Zhiling and Xie, Chenxuan and Cao, Meng and Gu, Yuxin and Hong, Sixin and Ren, Jing and Chen, Jian and Liu, Chao and Hua, Yining},
  year          = {2025},
  eprint        = {2504.19314},
  archivePrefix = {arXiv},
  primaryClass  = {cs.CL}
}

@misc{barres2025tau2benchevaluatingconversationalagents,
  title         = {{$\tau^2$-Bench}: Evaluating Conversational Agents in a Dual-Control Environment},
  author        = {Barres, Victor and Dong, Honghua and Ray, Soham and Si, Xujie and Narasimhan, Karthik},
  year          = {2025},
  eprint        = {2506.07982},
  archivePrefix = {arXiv},
  primaryClass  = {cs.AI}
}

@misc{ma2025korbenchbenchmarkinglanguagemodels,
  title         = {{KOR-Bench}: Benchmarking Language Models on Knowledge-Orthogonal Reasoning Tasks},
  author        = {Ma, Kaijing and Du, Xinrun and Wang, Yunran and Zhang, Haoran and Wen, Zhoufutu and Qu, Xingwei and Yang, Jian and Liu, Jiaheng and Liu, Minghao and Yue, Xiang and Huang, Wenhao and Zhang, Ge},
  year          = {2025},
  eprint        = {2410.06526},
  archivePrefix = {arXiv},
  primaryClass  = {cs.DB}
}

@inproceedings{shazeer2017moe,
  title     = {Outrageously Large Neural Networks: The Sparsely-Gated Mixture-of-Experts Layer},
  author    = {Shazeer, Noam and Mirhoseini, Azalia and Maziarz, Krzysztof and Davis, Andy and Le, Quoc V. and Hinton, Geoffrey E. and Dean, Jeff},
  booktitle = {International Conference on Learning Representations},
  year      = {2017}
}

@inproceedings{xiao2024streamingllm,
  title     = {Efficient Streaming Language Models with Attention Sinks},
  author    = {Xiao, Guangxuan and Tian, Yuandong and Chen, Beidi and Han, Song and Lewis, Mike},
  booktitle = {International Conference on Learning Representations},
  year      = {2024}
}
